\pdfoutput=1
\documentclass[11pt]{article}

\usepackage[preprint]{acl}

\usepackage{times}
\usepackage{latexsym}

\usepackage[T1]{fontenc}
\usepackage[utf8]{inputenc}

\usepackage{microtype}

\usepackage{inconsolata}

\usepackage{graphicx}

\newcommand{\PF}[1]{\raisebox{-.2em}{\includegraphics[height=1em]{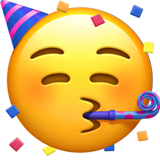}}}
\newcommand{\MF}[1]{\raisebox{-.2em}{\includegraphics[height=1.1em]{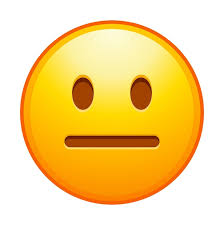}}}
\newcommand{\SF}[1]{\raisebox{-.2em}{\includegraphics[height=1em]{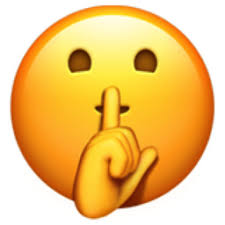}}}

\usepackage{comment}    % comment block
\usepackage{balance}
\usepackage{pifont}
\usepackage{amsmath}
\usepackage{amssymb}    % mathbb
\usepackage{diagbox}

\usepackage{fontawesome5}
\usepackage{longtable}

\usepackage{siunitx} % number alignment
\usepackage{booktabs} % Optional: Improves table aesthetics
\usepackage{multirow}
\usepackage{array}
\usepackage{rotating} % For rotated tables
\usepackage{adjustbox}
\usepackage{enumitem} % For compact contribution itemize
\usepackage{subcaption}
\usepackage{ulem}
\newcommand\highlight[1][yellow]{%
  \bgroup % 루아텍이면 주석처리
  \markoverwith{\textcolor{#1}{\vrule width.15em height.85em depth.2em}}%
  \ULon % 루아텍이면 주석처리
}

\newcommand{\hlhigh}[1]{\highlight[blue!20]{#1}}
\newcommand{\hlneu}[1]{\highlight[gray!30]{#1}}
\newcommand{\hllow}[1]{\highlight[red!10]{#1}}
\newcommand{\accatom}{$\text{ACC}_{\text{atom}}$}
\newcommand{\rcatom}{$\text{RC}_{\text{atom}}$}
\newcommand{\icatom}{$\text{IC}_{\text{atom}}$}

\title{Spotting Out-of-Character Behavior: \\Atomic-Level Evaluation of Persona Fidelity in Open-Ended Generation}

\newcommand\CoauthorMark{\footnotemark[\arabic{footnote}]}

\author{Jisu Shin\hspace{6mm}
    Juhyun Oh$\thanks{\hspace{2mm}Equally contributed.}$\hspace{6mm}
    Eunsu Kim$\CoauthorMark$\hspace{6mm}
    Hoyun Song\hspace{6mm}
    Alice Oh\\
    School of Computing\\
    Korea Advanced Institute of Science and Technology (KAIST)\\
    \texttt{\{jisu.shin,411juhyun,kes0317,hysong\}@kaist.ac.kr}
    \hspace{6mm}
    \texttt{alice.oh@kaist.edu}
    }

\begin{document}
\maketitle

\begin{abstract}
Ensuring persona fidelity in large language models (LLMs) is essential for maintaining coherent and engaging human-AI interactions. However, LLMs often exhibit Out-of-Character (OOC) behavior, where generated responses deviate from an assigned persona, leading to inconsistencies that affect model reliability. Existing evaluation methods typically assign single scores to entire responses, struggling to capture subtle persona misalignment, particularly in long-form text generation. To address this limitation, we propose an atomic-level evaluation framework that quantifies persona fidelity at a finer granularity. Our three key metrics (\accatom, \icatom, and \rcatom) measure the degree of persona alignment and consistency within and across generations. Our approach enables a more precise and realistic assessment of persona fidelity by identifying subtle deviations that real users would encounter. Through our experiments, we demonstrate that our framework effectively detects persona inconsistencies that prior methods overlook. By analyzing persona fidelity across diverse tasks and personality types, we reveal how task structure and persona desirability influence model adaptability, highlighting challenges in maintaining consistent persona expression.
\end{abstract}

\section{Introduction}

% 한국어: https://www.notion.so/loveydobby/250206-192f0f348a1b80b683dbeeb14af15776?pvs=4
% PPT: https://docs.google.com/presentation/d/1F4T2M9TcpiOtnKgmnYf88e1I_x6q0V1cFLuukokfcrU/edit?usp=sharing

\begin{table}[th!]
\small
\resizebox{\columnwidth}{!}{
\begin{tabular}{p{\columnwidth}}
    \toprule
    \multicolumn{1}{c}{\textbf{Instruction}}\\ 
    \midrule
    \texttt{You are \hlneu{neither extrovert nor introvert}.} \\
    Write an essay about today. \\ 
    \toprule
    \multicolumn{1}{c}{\textbf{Responses}}\\ 
    \midrule
    \textbf{\textit{Gen A}}:
      \hlneu{Today was just another day.} \hlneu{Had coffee in the morning, finished some work, and took a short walk.} \hlneu{Talked to a few people, nothing too exciting.} \hlneu{Spent the evening reading and watching a show.} \hlneu{Nothing special, nothing bad—just steady, like most days.} \hlneu{Maybe tomorrow will bring something different.} \\
    \midrule
    \textbf{\textit{Gen B}:}
      \hlhigh{Today was amazing! Morning coffee with friends was fun,} \hllow{but honestly, a bit draining.} 
      \hlhigh{Brainstorming was exciting—I love bouncing ideas around!}
      \hllow{Then suddenly, I just wanted to be alone. Read a book, felt peaceful.}
      \hllow{Later, friends called me out, and I went… but maybe I should’ve stayed home instead.} \\
    \toprule
    \multicolumn{1}{c}{\textbf{Overall Assessment}} \\
    \midrule
    \textbf{Prev.:} ``Both \textit{Gen A} and \textit{Gen B} are \MF{} on average.''\\
    \textbf{Ours:} ``\textit{Gen A} is \MF{} but \textit{Gen B} is mixed of \PF{} and \SF{}.''\\
    \bottomrule
\end{tabular}%
}
\centering 
\texttt{Persona} : \text{\texttt{{\hllow{Introvert}}(\SF{})-\hlneu{Neutral}(\MF{})-\hlhigh{Extrovert}(\PF{})}}
\caption{Generation examples from GPT-4o assigned ``\textit{neither extrovert nor introvert} (\MF{})'' persona. Previous evaluation treats both as equivalent, but ours distinguishes them, revealing Out-of-Character behavior.}
\label{tab:method_example}
\vspace{-0.2in}
\end{table}%

%\alice{the intro is a bit long, especially before you go to line 86, where you finally start to explain our work. this needs to come much earlier, preferably in the first page, but at least at the beginning of page 2.}
Assigning personas to large language models (LLMs) is essential for realistic and engaging human-AI interactions. 
The ability of LLMs to generate long-form text enables them to engage in complex interactions, such as dialogues~\cite{zhang-etal-2018-personalizing, lee-etal-2022-personachatgen}, role-playing~\cite{park2023generative, shao2023character}, narrative generation~\cite{malik2024empirical}, and social simulation~\cite{argyle2023out, ge2024scaling, hu-collier-2024-quantifying}.
However, ensuring consistent persona adherence in these complex interactions remains challenging, as even subtle inconsistencies can affect user perception and model reliability.
% particularly due to human sensitivity to even subtle inconsistencies.
% In this paper, we highlight the ``Out-of-Character (OOC)'' behavior in LLMs, where even minor deviations from the assigned persona can lead to user confusion, misinterpretation, and harm to the perceived reliability of the models.

% In the real world, when an individual or fictional character deviates from their established personality, they are considered to be acting OOC\footnote{For example, in \textit{<Harry Potter>}, Voldemort genuinely apologizing to humanity, or Hermione Granger skipping a class.}. These inconsistencies can disrupt the narrative and break the immersion in the character. This OOC behavior can also occur in persona-assigned LLMs.  %, as shown in Table~\ref{tab:method_example}.
% Table~\ref{tab:method_example} presents two different generations (Gen A and Gen B) from the LLM instructed to maintain a balanced persona (a person who is neither extrovert nor introvert).
% When we assign a balanced persona to an LLM, the desired outcome is a consistent portrayal of that persona, as Gen A exemplifies, where the language remains neither overly enthusiastic nor overly reserved.
% Contrary to this expectation, Gen B shows a mix of extroverted and introverted tendencies, which results in OOC behavior.
% Therefore, detecting OOC behavior is essential for ensuring high persona fidelity in LLMs.
One key challenge is Out-of-Character (OOC) behavior, where models deviate from their assigned persona. In the real world or fictional contexts, OOC behavior disrupts immersion, such as a well-known character acting against their established traits\footnote{For example, in \textit{<Harry Potter>}, Voldemort genuinely apologizing to humanity, or Hermione Granger skipping a class.}. Similarly, LLMs can inconsistently express personas, leading to confusion. \autoref{tab:method_example} illustrates this issue: while Gen A maintains a balanced persona (a person who is neither extrovert nor introvert), Gen B fluctuates between extroverted and introverted tendencies, demonstrating persona misalignment. Therefore, detecting such misalignment and inconsistencies is essential for ensuring high persona fidelity in LLMs.

Existing studies have focused on assessing persona fidelity at the response level, assigning a single score or representation to the entire generated text~\cite{wang-etal-2024-incharacter, wright-etal-2024-llm}.
While these approaches provide a general overview of persona adherence, they often struggle to capture the nuances of persona expression, especially in long-form text.
%As demonstrated in Table~\ref{tab:method_example}, previous methods assign similar scores to responses with vastly different levels of consistency, thus overlooking fluctuations in expressed personality.
%This limitation highlights the need for more fine-grained evaluation methods that can accurately assess persona adherence at a deeper level of granularity, capturing the dynamic interplay of personality traits within the text.
As shown in \autoref{tab:method_example}, different generations may receive similar scores under previous evaluations despite varying degrees of persona alignment and consistency, underscoring the need for a more fine-grained evaluation approach.
%%%%%%%%%%%%%%%%%%%%%%%%%%%%%%%%%%%%%%%%%

% To address this limitation, we propose an atomic-level evaluation framework for persona fidelity that enables a more precise and nuanced assessment of LLMs.
% Unlike previous approaches that rely on a single score for the entire response, our framework captures subtle persona misalignment and quantifies inconsistencies in the smaller textual units.
% This fine-grained analysis allows us to identify OOC behaviors that may be overlooked by previous methods, providing a deeper understanding of how well LLMs maintain their assigned personas throughout the generated text.
% By examining persona consistency at the atomic level, we can pinpoint the degree of deviation from the intended character, offering valuable insights for improving persona adherence and enhancing user experience.
To address this limitation, we propose an atomic-level evaluation framework that quantifies persona fidelity at a finer granularity. Unlike previous approaches, our framework identifies OOC behaviors at the atomic level, providing a more precise measure of persona alignment. By analyzing small textual units, we offer deeper insights into how well LLMs maintain their assigned personas throughout a response.

% 우리가 제안하는 메소드에 대한 구체적인 설명, 어떤 실험 했다.
% 또한 우리가 설정한 complex generation에서 발생할 수 있는 세 가지 주요 challenges에서의 persona fidelity를 측정하기 위해서 필요한 메트릭을 제시하였다.
% 우리 메트릭 세 개가 각각 뭘 재는지. + ooc
% 우리의 eval framework에서는 다측면의 persona fidelity를 측정하는 세 가지 메트릭을 제시한다.
% \accatom은 모델이 주어진 페르소나에 align하는 생성 능력을 측정하여 glitch---설명필요/튀는 문장---의 비율?을 알 수 있다. at the atomic level. \icatom은 생성문 내의 페르소나 특성의 변동성을 측정하여 모델의 담화가 얼마나 일관된지를 암시한다. 마지막으로 \rcatom은 생성문이 반복되어도 그 atomic score의 분포가 일관되게 생성되는지를 측정하여 반복에 따른 변덕 문제를 캐치할 수 있다.
% 이 워크에서는 실험을 통해 우리의 평가 방법과 제시하는 메트릭은 이전의 메트릭보다 더 상세한 레벨의 persona misalignment를 발견할 수 있음을 보였다. 또한 우리는 모델의 다양한 생성 태스크와 페르소나에서의 페르소나 충실도를 평가하여 페르소나 부여 모델의 다양한 조건에서의 interactive behavior?을 분석하였다. 우리의 컨트리뷰션은 다음과 같다.
% 이 문단에 성격 가지고 실험한다고 추가
% To this end, we introduce atomic-level evaluation with three key metrics to assess multiple aspects of persona fidelity.
% \accatom\ measures a model’s ability to generate text aligned with the assigned persona, capturing the proportion of off-character sentences at the atomic level.
% \icatom\ quantifies variability in persona characteristics within a single response, providing insights into the internal coherence of a model’s discourse.
% \rcatom\ evaluates the consistency of atomic-level score distributions across repeated generations, detecting fluctuations in persona expression over multiple responses.
%To this end, we introduce atomic-level evaluation with three key metrics to assess multiple aspects of persona fidelity.
We introduce three key metrics to assess multiple aspects of persona fidelity.
\accatom\ measures persona alignment, detecting off-character sentences within a response.
\icatom\ and \rcatom\ assess intra- and inter-generation consistency, capturing fluctuations in persona expression. %by capturing variability in persona expression within a single generation.
% \rcatom evaluates reproducibility, detecting fluctuations in persona expression across repeated generations.
Our study focuses on personality-based personas, examining how well models maintain assigned personality traits in diverse open-ended generation tasks.
Through experiments, we demonstrate that our framework identifies subtle persona misalignment better than previous methods. By analyzing persona fidelity across different tasks and persona types, we underscore the behavior of persona-assigned models under varying conditions.

Our contributions are as follows:
\begin{itemize}[leftmargin=*,topsep=1px,partopsep=0px]
\setlength\itemsep{-0.4em}
    \item This study is the first to explore subtle OOC behavior in persona-assigned LLMs during open-ended text generation, where deviations from the expected persona can undermine user trust and model reliability.
    \item We propose a fine-grained evaluation framework with three atomic-level metrics---\accatom, \icatom, and \rcatom---to capture subtle persona misalignment and enable a more precise measurement of persona fidelity.
    \item Our experiments demonstrate the effectiveness of our framework in assessing persona fidelity across diverse conditions, emphasizing the challenges in maintaining persona fidelity in each of these contexts.

\end{itemize}

\section{Related Work}
\label{sec:related_work}
% https://www.notion.so/loveydobby/241204-151f0f348a1b8002b464fc2de66209b4?pvs=4

\paragraph{Persona-Assigned LLM}
%persona assigning
%Large language models (LLMs) have been studied for their opinions on subjective tasks, including personality tests~\cite{safdari2023personality, jiang2024evaluating}, value assessments~\cite{miotto-etal-2022-gpt}, political compass tests~\cite{feng-etal-2023-pretraining}, and moral choices~\cite{rao-etal-2023-ethical, scherrer2024evaluating}.
%These tasks lack definitive or universally correct answers, as responses can vary depending on the individual, and LLMs are not designed to represent any specific person.
%However, there are preferred or desirable answers that are expected from steered LLMs~\cite{benkler2023assessing, opinionqa-kind}.
% For example, a user who interacts with a chatbot whose persona has the same political orientation as the user will expect to discuss a political issue from the same point of view.
% Therefore, LLMs are evaluated on the fidelity~\cite{argyle2023out}---how their responses are aligned with the given personas or steered characters.
% steerability와 consistency를 기준으로 제시하여 둘을 측정했다.
The effects of persona on LLMs have been studied in many tasks, such as dialogue~\cite{zhang-etal-2018-personalizing, wan2023personalized}, reasoning~\cite{kong-etal-2024-better, gupta2024bias, salewski2024context}, and LLM safety tasks~\cite{deshpande-etal-2023-toxicity, ko2024different}.
For subjective tasks, the persona-assigned LLMs have been analyzed to identify their opinions on personality~\cite{safdari2023personality, jiang2024evaluating}, social value~\cite{miotto-etal-2022-gpt, durmus2023towards}, political orientation~\cite{feng-etal-2023-pretraining, liu-etal-2024-evaluating-large}, and moral decision-making~\cite{benkler2023assessing, rao-etal-2023-ethical, scherrer2024evaluating}, as well as to investigate which groups these opinions aligned with~\cite{santurkar-2023-whose, hwang-etal-2023-aligning, sun2023aligning}.
Persona alignment in subjective tasks is crucial for pluralism, enabling models to represent and speak for diverse individuals fairly and inclusively~\cite{sorensen2024position}.

\paragraph{Persona Fidelity Evaluation}
%  consistency도 measure해야 한다, 
The desire to ensure that models assigned a persona accurately reproduce and simulate the given persona has driven research into persona fidelity measurement~\cite{argyle2023out}.
% persona assigned model이 입력된 persona를 그대로 재현/시뮬레이트 하기를 원하는 사람들의 기대는 persona fidelity measurement 연구로 발전함.
% fidelity는 steerability를 측정하려는 여러 시도들이 있었고,  그러나 LLMs are evaluated on the fidelity~\cite{argyle2023out}---how their responses are aligned with the given personas or steered characters.
While early evaluations focused on whether the average measured personality traits aligned with the given persona~\cite{jiang2024evaluating}, more recent work has expanded to assess the consistency of measured personality across various settings~\cite{shu-etal-2024-dont, wang-etal-2024-incharacter}.
% Persona fidelity의 eval은 measure된 성격점수의 평균적인 경향이 given persona와 align이 되는가~\cite{jiang2024evaluating}에서 더 나아가, measured personality가 across various settings에서 얼마나 consistent한지까지 함께 평가되고 있다~\cite{shu-etal-2024-dont, wang-etal-2024-incharacter}. 
Previous research has often identified inconsistencies in large language models (LLMs) when exposed to spurious variations in prompt format (e.g., colons, sentence endings, order bias)~\cite{shu-etal-2024-dont, gupta2024bias, huang-etal-2024-reliability} or contentual variations based on paraphrasing (e.g., negation)~\cite{dorner2023personality, pellert2024ai}.
% 기존 연구들은 prompt format의 spurious variations (e.g., colons, sentence ending, order bias) 관점~\cite{shu-etal-2024-dont, gupta2024bias, huang-etal-2024-reliability} 혹은 paraphrase 기반의 contentual variation (e.g., negation)~\cite{}의 관점에서 llm이 consistency가 떨어진다는 것을 발견했다. 
However, these studies primarily relied on multiple-choice questions or closed-ended questions (e.g., yes/no, rating scale responses). Given that response tendencies can vary based on question format~\cite{west2023generative, rottger2024political, liu-etal-2024-evaluating-large, wright-etal-2024-llm}, it is uncertain whether findings from multiple-choice question or short-form evaluation generalize to generation tasks. For this reason, we evaluate persona fidelity in a generation setting.
% 그러나 이런 연구들은 대부분 mcq나 closed-ended question (예/아니오, 점수로 답을 내는)에서 진행되었다. 질문 방식에 따라 모델의 응답 경향성이 다르기 때문에~\cite{west2023generative} persona test를 mcq, short-form로 물어봤을 때의 결과가 generation에도 동일하게 드러날지 확신할 수 없다. 이런 이유로 우리는 generation setting에서 persona fidelity를 측정한다.

% 응답의 평균 경향성(mean)이 persona와 일치하는지를 보는 steerability와 여러 번의 응답에서 일관되게/robust하게 persona를 지킬 수 있는지를 보는 consistency를 기준으로 제시하여 둘을 측정했다.
% 선행연구들은 MCQ, closed-form에서 평가했다.
% 선행연구는 페르소나 llm이 mcq 셋팅에서 일관성을 보이지 못한다는 점을 발견했다.
% 기존의 연구는 mcq 환경에서 드러나는 low confid ence on prompt variation and content-level variation.
% 1안 - 또한 질문 방식에 따라 모델의 응답 경향성이 다르기 때문에 questionnaire task를 mcq로 물어봤을 때의 결과가 generation에도 동일하게 드러날지 확신할 수 없다. 이런 이유로 우리는 generation setting에서 persona fidelity를 측정한다.

\begin{figure*}[th!]
    \centering
    \includegraphics[width=\linewidth]{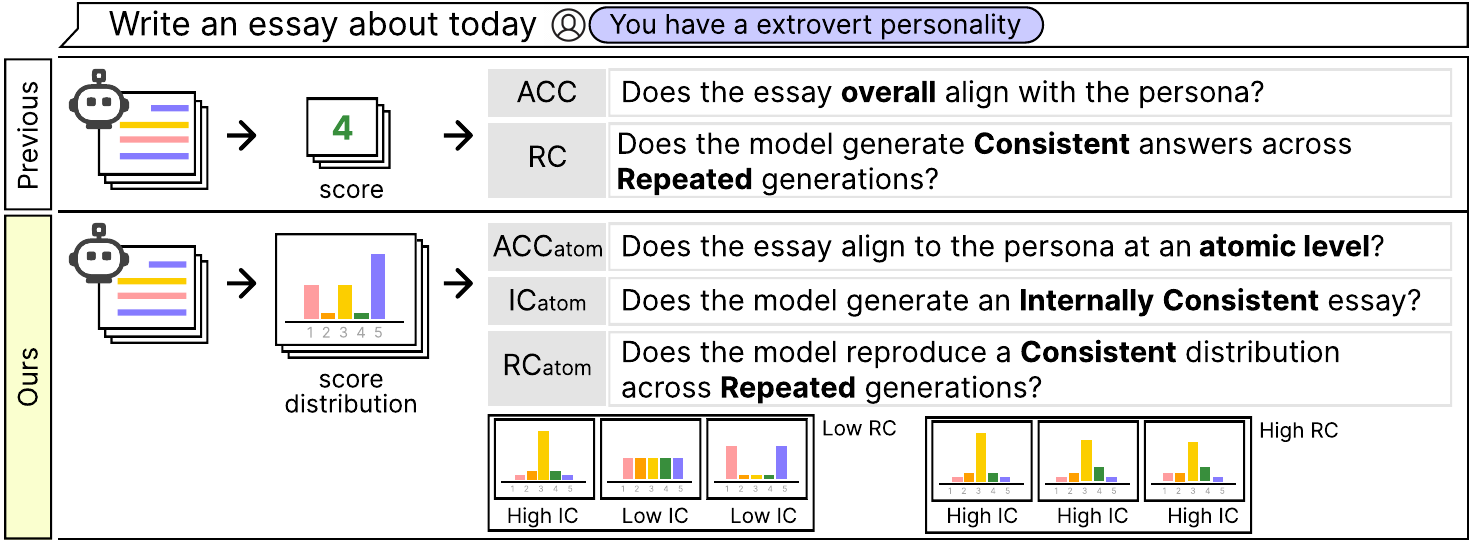}
    \caption{Overview of our evaluation method and proposed metrics. A previous method assigns a single score to a long-form generation, collapsing multiple persona-related information within the generation. In contrast, our approach evaluates atomic units, preserving each piece of information and allowing for a more fine-grained assessment of persona fidelity.}
    \label{fig:method_overview}
    \vspace{-0.2in}
\end{figure*}

\paragraph{Persona Fidelity Evaluation in Open-Ended Generation Tasks}
Existing research on persona fidelity has primarily focused on closed-form generation tasks, employing diverse evaluation methodologies. Approaches include human annotations~\cite{jiang2024evaluating}, statistical analyses of linguistic features~\cite{safdari2023personality, argyle2023out, jiang-etal-2024-personallm}, and model-based scoring techniques that feed generated responses back into language models to extract characteristic assessments~\cite{liu-etal-2024-evaluating-large, wang-etal-2024-incharacter, jiang-etal-2024-personallm}.
% However, these methods typically reduce a complex generation to a single score, neglecting the nuanced challenges and subtle OOC problems of a long-form, open-ended generation.
% Transformer-based language models are known to struggle with maintaining coherence in extended outputs~\cite{sun-etal-2021-long, deng2022model, krishna-etal-2022-rankgen}. Our work addresses this limitation by introducing a novel approach that evaluates inner-consistency through atomic facts.
However, these methods typically reduce a complex generation to a single score, neglecting the nuanced challenges and subtle OOC problems of a long-form, open-ended generation. This limitation makes it challenging to identify the difficulties transformer-based language models face in maintaining coherence in extended outputs~\cite{sun-etal-2021-long, deng2022model, krishna-etal-2022-rankgen}. Our work addresses this limitation by introducing a novel approach that evaluates internal consistency through atomic facts.

\section{Atomic-Level Evaluation for Persona Fidelity}
\label{section:method}

We propose a new evaluation framework that measures persona fidelity at the \textbf{atomic level}, capturing the multifaceted nature of persona expression in long-form generations. These generations comprise multiple sentences and expressions, each contributing to the overall persona perception. Analyzing persona alignment within smaller textual units allows us to detect nuanced variations and inconsistencies often overlooked by previous methods.

%\subsection{Framework for Atomic-Level Persona Fidelity Evaluation}
%\label{section:method-subsection:framework}
% 어쩌구저쩌구~
% 이런 걸 잴 거고요
% A-ACC (atomic-level persona accuracy)
% ATRT (consistency of atomic sentence distribution on repeated trials)
% AIC
% TC
% 이렇게 잴 거예요
% 그래서 테이블 1의 문장을 우리 걸로 이렇게 해석할 거예요.

To evaluate atomic-level persona fidelity in models' responses, we introduce the three metrics---\accatom, \icatom, and \rcatom---that provide a comprehensive understanding of a model’s ability to embody and maintain assigned personas.
Motivated by FActScore~\cite{min-etal-2023-factscore}, we evaluate persona fidelity by aggregating the characteristic scores of atomic units.
In prior studies~\cite{nenkova-passonneau-2004-evaluating, shapira-etal-2019-crowdsourcing, zhang-bansal-2021-finding, liu-etal-2023-revisiting}, atomic units are defined as the smallest units of statements that contain information. In our study, we define atomic units as the smallest textual segments that convey persona-relevant characteristics.
Thus, a generated response $G$ is divided into atomic units $\{a_1, ..., a_n\}$, where each unit $a_i$ is assigned a characteristic score $s_i$ that represents the level of a specific personality trait or value dimension\footnote{For example, in the agreeableness personality dimension, $a_i$ receives the maximum score for $s_i$ when it expresses highly agreeable content; conversely, it receives a low score when it conveys disagreeable or unfriendly content.}. These scores are determined either using a scoring model or predefined criteria.
 
% introduction
\paragraph{Atomic-level Accuracy (\accatom)} measures the degree to which individual atomic units align with an assigned persona. Unlike previous persona evaluation methods that assess fidelity at the response level, \accatom\ evaluates persona alignment at a finer granularity.
The \accatom\ score for a response $G$ is then computed as the mean accuracy score across all atomic units---specifically, whether the assigned characteristic scores match the target score:
\begin{equation*}%\[
\text{ACC}_{\text{atom}} = \frac{1}{n} \sum_{i=1}^{n} \mathbb{I}(s_i = s_{target} )
\end{equation*}%\]
, where \( n \) is the number of atomic units in the response, and $\mathbb{I}(\cdot)$ is the indicator function which returns 1 if $s_i$ matches the target score $s_{target}$, and 0 otherwise.

\paragraph{Atomic-level Internal Consistency (\icatom)} evaluates the consistency of persona expression within a single generated response.
With the shift from multiple-choice and single-score assessments to open-ended generation, internal consistency becomes a crucial evaluation aspect, as it reflects coherence and stability in longer responses.
For each generated response $G$, we represent its atomic characteristic scores as a frequency distribution $D$, where each bin corresponds to a characteristic score and its frequency represents the number of atomic units assigned that score.
\icatom\ is computed as the inverse of the normalized standard deviation of the characteristic score distribution $D$:
\begin{equation*}%\[
\text{IC}_{\text{atom}} = 1 - \text{STD}(D) \times \frac{2}{\max-\min}
\end{equation*}%\[
, where $\text{STD}(D)$ is the standard deviation of the characteristic scores within $G$, and $\max$ and $\min$ represent the maximum and minimum characteristic scores.
For interpretability, we normalize the standard deviation to a [0,1] range and apply an inverse transformation so that higher \icatom\ scores indicate greater internal consistency, while lower scores suggest misalignment or fluctuation in persona fidelity within a single response.

\paragraph{Atomic-level Retest Consistency (\rcatom)} assesses the reproducibility of persona alignment across repeated generations for the same input.
%Reproductive Consistency (RC), analogous to test-retest reliability in psychometrics~\cite{guttman1945basis}, measures whether a model generates the same responses when prompted multiple times under identical conditions.
Analogous to test-retest reliability in psychometrics~\cite{guttman1945basis}, RC measures whether a model consistently generates persona-aligned responses when prompted multiple times under identical conditions.
% Unlike prior studies that evaluated consistency using methods such as T-tests~\cite{huang-etal-2024-reliability} or standard deviation~\cite{wang-etal-2024-incharacter}, 
Unlike prior studies that measure RC with the variance of scores at the response level~\cite{huang-etal-2024-reliability, wang-etal-2024-incharacter}, we adopt the Earth Mover's Distance (EMD)~\cite{rubner1998metric} to measure distributional differences between repeated generations.
The EMD quantifies the minimum cost required to transform one score distribution into another, offering a more comprehensive evaluation of consistency at the distribution level.
Given multiple generations for the same prompt, we compute the EMD between every unique pair of distributions ($D_i$, $D_j$), ensuring that each pair satisfies $i \neq j$ and is counted only once (i.e., $i<j$).
The final \rcatom\ score is obtained by averaging the EMD scores across all unique pairs and applying a normalization function to ensure values remain within a standardized range:
\begin{equation*}
\text{RC}_{\text{atom}} = \text{normalize} \left( \frac{1}{|P|} \sum_{(i,j) \in P} EMD(D_i, D_j) \right)
\end{equation*}%
, where $P$ is the set of all unique pairs of generations for a given prompt, and \( \text{normalize}(x) \) is defined as:
\(
\text{normalize}(x) = \left(1 - \frac{x}{\max - \min} \right) \times 2 - 1
\)
.
This normalization function scales the EMD values to a [-1,1] range, where a higher \rcatom\ score indicates greater distributional consistency in persona fidelity across repeated generations, while a lower score reflects more variability.

\section{Experimental Setup}
\label{section:experimental_setup}

\subsection{Tasks}
\label{section:experimental_setup-tasks}

\paragraph{Personality Domain}
Personality is a fundamental characteristic that defines individual differences.
It serves as a key domain to assess whether LLMs can effectively and faithfully align their responses with the fundamental characteristics of human personality in role-playing and persona alignment research~\cite{safdari2023personality, huang-etal-2024-reliability, jiang-etal-2024-personallm}.
In this study, we focus on personality personas, selecting them as our primary domain among various possible persona types.
To examine this, we adopt the Big 5 Personality Traits~\cite{goldberg1992big5}, which defines personality along five dimensions (OCEAN): \textbf{O}penness to experience, \textbf{C}onscientiousness, \textbf{E}xtraversion, \textbf{A}greeableness, and Emotional Stability (\textbf{N}euroticism).

\paragraph{Generation Tasks}
We evaluate persona fidelity across three open-ended generation tasks, specifically designed to assess whether LLMs generate responses aligned with their assigned personality personas.
\textbf{1) Interview with Questionnaires}
The most widely used method for evaluating LLMs’ personality fidelity is self-report questionnaires. Following previous studies~\cite{trull1998structured, wang-etal-2024-incharacter}, we convert multiple-choice questionnaire items into open-ended interview-style questions (e.g., transforming “\textit{I have a rich vocabulary. (A) Very inaccurate ($\ldots$) (E) Very accurate.}” into “\textit{Do you have a rich vocabulary?}”). We employ IPIP-BFFM~\cite{goldberg1992big5}, which consists of 10 questions per personality dimension. Each model is assigned questions corresponding only to the personality dimension of its assigned persona (e.g., a model assigned the Openness persona responds solely to the questions of the Openness dimension).
\textbf{2) Essay}
% The essay writing task proposed by \citet{kwantes2016assessing} is designed to predict personality traits based on language use.
% In this task, LLMs are provided with a background context before writing an essay about their thoughts and perspectives within that given situation~\cite{jiang2024evaluating}.
% %NLP연구에서는 \cite{jiang2024evaluating}이 essay writing에 성격 페르소나가 영향을 미치는 것을 확인하였다.
% Similar to the questionnaire task, we provide the essay context that corresponds to the same personality dimension as the assigned persona. 
% %persona에 해당하는 essay task (given context) 준다.
The essay writing task~\cite{kwantes2016assessing} is commonly used to assess personality traits based on linguistic expression. In recent research, \citet{jiang2024evaluating} demonstrated that personality personas influence LLMs' essay writing. In this task, LLMs are provided with a background context (scenario) and generate an essay reflecting their thoughts and perspectives. Similar to the questionnaire task, models generate responses to scenario prompts corresponding to their assigned persona’s personality dimension.
\textbf{3) Social Media Post}
% 제안 \cite{park2015automatic} 얘네가 personality가 language use에서 많이 드러난다는 걸 주장하기 위해서 facebook status를 작성한 걸 기반으로 성격을 오토매틱하게 분석함.
% \cite{safdari2023personality} llm에 사용해서 persona가 social meadia post 에서 controlled generation의 condition으로 사용될 수 있음을 확인함.
\citet{park2015automatic} demonstrated that personality traits are reflected in social media language through the automatic analysis of Facebook status updates.
Building on this, \citet{safdari2023personality} demonstrated that LLMs could incorporate personality personas as controlled generation conditions in social media post writing. We prompt models to generate social media posts and assess whether their persona alignment is maintained in free-form content.
Further details on task prompts are provided in Appendix~\ref{appendix:setup-tasks}.

\subsection{Details for Setup}
\label{section:experimental_setup-details}

\begin{table}[t!]
    \centering
    \resizebox{\linewidth}{!}{
    \begin{tabular}{c|c|c|c}
        \toprule
         & \textbf{High} & \textbf{Neutral} & \textbf{Low}\\
         \midrule
        \textbf{O} & open & neither open nor close-minded & close-minded\\
        \textbf{C} & conscientious & neither conscientious nor careless & careless\\
        \textbf{E} & extroverted & neither extroverted nor introverted & introverted \\
        \textbf{A} & agreeable & neither agreeable nor disagreeable & disagreeable \\
        \textbf{N} & emotionally stable & neither emotionally stable nor neurotic & neurotic \\   
        \bottomrule
    \end{tabular}
    }
    \caption{Personality traits used for our persona entities.}
    \label{tab:persona_list}
    % \vspace{-0.2in}
\end{table}
\paragraph{Personas}
We define 15 personality personas, each corresponding to a high, neutral, or low\footnote{Note that `high,' `neutral,' and `low' do not imply positive or negative traits, but rather indicate relative positions on the personality trait scale.} score on one of the five personality dimensions (\autoref{tab:persona_list}).
For instance, in the \textbf{E}xtraversion dimension, the high-level persona is labeled as ``\textit{extroverted person},'' the neutral-level persona as ``\textit{neither extroverted nor introverted person},'' and the low-level persona as ``\textit{introverted person}.''
% Each model is assigned a persona by prompting it with one of six persona-assignment instructions, and details of persona descriptions and assignment prompts can be found in Appendix~\ref{appendix:persona}.
Due to the page limit, we provide six persona-assignment instructions in Appendix~\ref{appendix:setup-persona}.

\paragraph{Models}
We evaluate persona fidelity across 12 widely used LLMs, categorized into base models and tuned models.
For the base models, we utilize Davinci-002 from OpenAI, LLaMA-3-8B from Meta, and Mistral-7B-v0.3 from Mistral AI.
For the tuned models, we use four GPT models (GPT-3.5-turbo~\cite{ouyang2022training}, GPT-4-turbo~\cite{achiam2023gpt4}, GPT-4o, and GPT-4o-mini), two LLaMA models (LLaMA-3-8B-Instruct and LLaMA-3-70B-Instruct~\cite{dubey2024llama3}), one Mistral model (Mistral-7B-instruct-v0.3~\cite{jiang2023mistral}) and two Claude models (Claude-3-haiku and Claude-3-sonnet).
All models are evaluated across 30 runs per task and persona.
Further details on prompt settings and hyperparameters for each model are provided in Appendix~\ref{appendix:setup-model}.

\paragraph{Scoring and Filtering}
We set a sentence as an atomic unit, and divide a long generation into atomic units using \texttt{sent\_tokenize} function from NLTK.
To automatically convert an atomic unit into a characteristic score~\cite{wang-etal-2024-incharacter}, we utilize GPT-4o as the scoring model.
Each atomic sentence in a generated response is assigned a personality score within the range [1,5], where 5 represents a high level of the corresponding personality trait\footnote{e.g., In the Openness dimension: ``1: Very close-minded -- 5: Very open-minded''.}.
Our experimental results report persona fidelity only for valid sentences after filtering invalid sentences where personality scores could not be assigned (e.g., ``\textit{What a whirlwind few weeks it's been!}'').
See Appendix~\ref{appendix:setup-scoring model} for more details.

\paragraph{Metrics}
In our experiment, we compare three proposed metrics to two previous metrics.
As described in Section~\ref{section:method}, our metrics---$\text{ACC}_{\text{atom}}$, $\text{RC}_{\text{atom}}$, and $\text{IC}_{\text{atom}}$---capture atomic-level persona alignment and multi-dimensional consistencies.
For comparison, we evaluate two previously established metrics, ACC and RC\footnote{ACC and RC correspond to $\text{\textbf{ACC}}_{\texttt{Dim}}$ and $\text{\textbf{Std}}_{\texttt{Score}}$ metrics of \citet{wang-etal-2024-incharacter}, respectively.}, which are computed based on overall response-level scores.
The ACC score for a response $G$ is computed as follows:
\(
\text{ACC} = \mathbb{I}(s=s_{target})
\)
, where $s$ means a single overall characteristic score.
RC is calculated as the standard deviation of overall scores, providing insight into reproducibility across repeated generation:
\(
\text{RC} = 1-\frac{2 \times \text{STD}(\{s_{G_1}, ..., s_{G_n}\})}{\max - \min}
\).
For \accatom\ and ACC, we divide the [1,5] score range into three equal sections: high, neutral, and low\footnote{The target score ranges ($s_{target}$) are [1, 2.33) for low-level personas, [2.33,3.67) for neutral-level personas, and [3.67, 5] for high-level personas.}. A response is considered persona-aligned if the assigned persona level matches the range in which its characteristic score falls.

\begin{table}[t!]
    \centering
    \resizebox{\linewidth}{!}{
    \begin{tabular}{l|lllll}
        \toprule
         & \multicolumn{1}{c}{O} & \multicolumn{1}{c}{C} & \multicolumn{1}{c}{E} & \multicolumn{1}{c}{A} & \multicolumn{1}{c}{N} \\
        \midrule
        Kendall's $\tau$  & 0.69$^{***}$ & 0.76$^{***}$ & 0.67$^{***}$ & 0.72$^{***}$ & 0.69$^{***}$ \\
        % Significance     & *** & *** & *** & *** & *** \\
        \midrule
        Fleiss' $\kappa$ & 0.90 & 0.96 & 0.80 & 0.84 & 0.74\\
        \bottomrule
        \multicolumn{6}{l}{\small ***: $p<.001$}
    \end{tabular}
    }
    \caption{Results of the human evaluation for LLM-based scoring across the five personality dimensions. Kendall's $\tau$ indicates the rank correlation between GPT-4o scores and human judgments; Fleiss' $\kappa$ reflects inter-annotator agreement.}
    \label{tab:human_evaluation}
\end{table}
\subsection{Human Validation}

To validate the reliability of our LLM-based scoring, we conduct a human evaluation for each of the five personality traits, using 50 sentence pairs per trait (250 pairs in total).
Each pair includes two atomic sentences with differing LLM-based scores, and six annotators rank which sentence conveys a higher level of the target personality trait.
% Before presenting the main experimental results, we validate the reliability of our LLM-based evaluation framework through a human evaluation.
As shown in \autoref{tab:human_evaluation}, the results demonstrate strong alignment between human judgments and model scores, with Kendall's $\tau$~\cite{kendall1938new} ranging from 0.67 to 0.76 (all $p<.001$), indicating high correlation.
Inter-annotator agreement was also high, with Fleiss' $\kappa$~\cite{fleiss1971measuring} values between 0.74 and 0.96, supporting the validity of our automatic scoring method.
We provide details for human evaluation in Appendix~\ref{appendix:setup-human validation}.

\section{Experimental Results}
\label{section:results}

% subsection 1

\begin{table}[t]
    \centering
    %\resizebox{0.7\linewidth}{!}{
    \small
    \begin{tabular}{cc|ccc}
    \toprule
    & & \multicolumn{3}{c}{Ours} \\ 
    & & \accatom & \rcatom & \icatom \\
    \midrule
        \multirow{2}{*}{Prev.} & ACC & 0.91	& 0.51	& \textbf{0.40} \\
        & RC & 0.48	& 0.98	& \textbf{0.37} \\
    \bottomrule
    \end{tabular}
    %}
    \caption{Pearson correlation coefficient ($r$) between previous metrics (Prev.) and atomic-level metrics (Ours). %The high values of $r_{\text{ACC}, \text{ACC}_{\text{atom}}}$ and $r_{\text{RC}, \text{RC}_{\text{atom}}}$ suggests that our metrics captures accuracy and reproducibility similarly to traditional ones, while \icatom\ provides complementary insights.}
    }
    \label{tab:correlation_metrics_old_ours}
    \vspace{-0.1in}
\end{table}
\begin{figure}[t]
    \centering
    \includegraphics[width=1\linewidth]{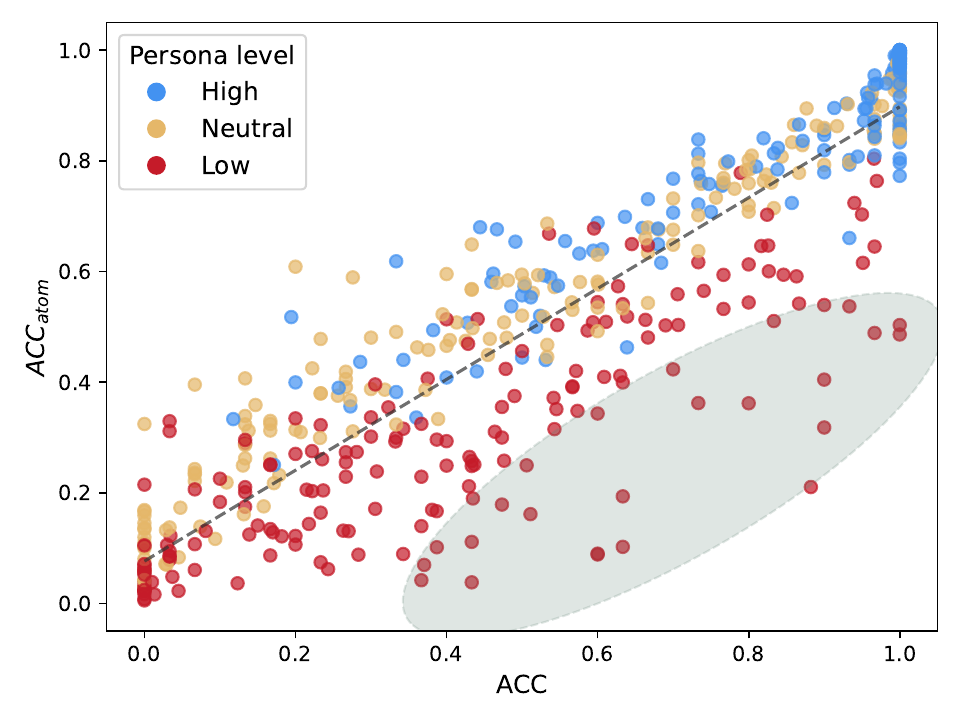}
    \caption{
    The relationship between ACC and \accatom\ of all persona-assigned models. The shaded region highlights instances that do not follow the correlation.
    }
    \label{fig:comparison_prev_ours}
    \vspace{-0.1in}
\end{figure}
% subsection 2
\begin{table*}[t!]

\resizebox{\textwidth}{!}{%
\begin{tabular}{ccc|cccc|cccc|cccc}

\toprule

 &
 &
  \multicolumn{1}{r|}{Tasks} &
  \multicolumn{4}{c|}{Interview with Questionnaires} &
  \multicolumn{4}{c|}{Essay} &
  \multicolumn{4}{c}{Social Media Post}
   \\

\midrule

\multicolumn{1}{l}{} &
  \multicolumn{1}{l}{} &
  \multirow{2}{*}{\diagbox[width=2.8cm, height=1cm]{Dimension}{Metrics}} &
  \multicolumn{1}{c}{\small MEAN} &
  \multicolumn{1}{c}{\small \accatom} &
  \multicolumn{1}{c}{\icatom} &
  \multicolumn{1}{c|}{\rcatom} &
  
  \multicolumn{1}{c}{\small MEAN} &
  \multicolumn{1}{c}{\small \accatom} &
  \multicolumn{1}{c}{\icatom} &
  \multicolumn{1}{c|}{\rcatom} &
  
  \multicolumn{1}{c}{\small MEAN} &
  \multicolumn{1}{c}{\small \accatom} &
  \multicolumn{1}{c}{\icatom} &
  \multicolumn{1}{c}{\rcatom} \\
  
  &
  Level &
   &
  \multicolumn{1}{c}{[1,5]} &
  \multicolumn{1}{c}{[0,1]} &
  \multicolumn{1}{c}{[0,1]} &
  \multicolumn{1}{c|}{[-1,1]} &
  
  \multicolumn{1}{c}{[1,5]} &
  \multicolumn{1}{c}{[0,1]} &
  \multicolumn{1}{c}{[0,1]} &
  \multicolumn{1}{c|}{[-1,1]} &
  
  \multicolumn{1}{c}{[1,5]} &
  \multicolumn{1}{c}{[0,1]} &
  \multicolumn{1}{c}{[0,1]} &
  \multicolumn{1}{c}{[-1,1]} \\

\midrule

 &
   &
  O &
  4.60 &
  \cellcolor[HTML]{3672DA}0.94 &
  \cellcolor[HTML]{8CB978}0.82 &
  \cellcolor[HTML]{EBA65E}0.87 &
  4.66 &
  \cellcolor[HTML]{2C6AD8}0.99 &
  \cellcolor[HTML]{A4C895}0.76 &
  \cellcolor[HTML]{E99B4A}0.93 &
  4.71 &
  \cellcolor[HTML]{2E6CD8}0.98 &
  \cellcolor[HTML]{A4C895}0.76 &
  \cellcolor[HTML]{E99D4D}0.92 
\\
 &
   &
  C &
  4.83 &
  \cellcolor[HTML]{2E6CD8}0.98 &
  \cellcolor[HTML]{6FA756}0.89 &
  \cellcolor[HTML]{E99D4D}0.92 &
  4.93 &
  \cellcolor[HTML]{2C6AD8}0.99 &
  \cellcolor[HTML]{66A24D}0.91 &
  \cellcolor[HTML]{E89743}0.95 &
  4.64 &
  \cellcolor[HTML]{2E6CD8}0.98 &
  \cellcolor[HTML]{A4C895}0.76 &
  \cellcolor[HTML]{EAA154}0.90 
\\
 &
   &
  E &
  4.90 &
  \cellcolor[HTML]{2C6AD8}0.99 &
  \cellcolor[HTML]{66A24D}0.91 &
  \cellcolor[HTML]{E89946}0.94 &
  4.03 &
  \cellcolor[HTML]{467CDD}0.87 &
  \cellcolor[HTML]{C1D9B7}0.69 &
  \cellcolor[HTML]{EDB06F}0.82 &
  3.88 &
  \cellcolor[HTML]{5789E0}0.79 &
  \cellcolor[HTML]{ADCD9F}0.74 &
  \cellcolor[HTML]{EAA154}0.90 \\
 &
   &
  A &
  4.89 &
  \cellcolor[HTML]{306DD9}0.97 &
  \cellcolor[HTML]{629F48}0.92 &
  \cellcolor[HTML]{E99D4D}0.92 &
  4.52 &
  \cellcolor[HTML]{487EDD}0.86 &
  \cellcolor[HTML]{C6DCBC}0.68 &
  \cellcolor[HTML]{EDAE6C}0.83 &
  4.94 &
  \cellcolor[HTML]{2C6AD8}0.99 &
  \cellcolor[HTML]{6AA451}0.90 &
  \cellcolor[HTML]{E7933C}0.97 \\
 &
   &
  N &
  4.97 &
  \cellcolor[HTML]{2968D7}1.00 &
  \cellcolor[HTML]{4D922F}0.97 &
  \cellcolor[HTML]{E69138}0.98 &
  4.39 &
  \cellcolor[HTML]{326FD9}0.96 &
  \cellcolor[HTML]{B9D4AD}0.71 &
  \cellcolor[HTML]{EDAE6C}0.83 &
  4.95 &
  \cellcolor[HTML]{2968D7}1.00 &
  \cellcolor[HTML]{66A24D}0.91 &
  \cellcolor[HTML]{E7933C}0.97  \\

\cmidrule{3-15}

 &
  \multirow{-6}{*}{High} &
  Total &
  4.84 &
  0.98 &
  0.90 &
  0.93 &
  4.51 &
  0.93 &
  0.75 &
  0.87 &
  4.62 &
  0.95 &
  0.81 &
  0.93  \\

\cmidrule{2-15}

 &
   &
  O &
  \textit{\textbf{4.03}} &
  \cellcolor[HTML]{E3ECFA}0.14 &
  \cellcolor[HTML]{9CC38B}0.78 &
  \cellcolor[HTML]{EBA65E}0.87 &
  \textit{\textbf{4.58}} &
  \cellcolor[HTML]{FFFFFF}0.01 &
  \cellcolor[HTML]{A0C590}0.77 &
  \cellcolor[HTML]{EAA357}0.89 &
  \textit{\textbf{4.73}} &
  \cellcolor[HTML]{FBFCFF}0.03 &
  \cellcolor[HTML]{A0C590}0.77 &
  \cellcolor[HTML]{E99D4D}0.92  \\
 &
   &
  C &
  3.59 &
  \cellcolor[HTML]{A7C1EF}0.42 &
  \cellcolor[HTML]{98C087}0.79 &
  \cellcolor[HTML]{F0B981}0.77 &
  \textit{\textbf{4.02}} &
  \cellcolor[HTML]{C7D8F5}0.27 &
  \cellcolor[HTML]{CEE1C5}0.66 &
  \cellcolor[HTML]{F4CCA3}0.67 &
  \textit{\textbf{4.22}} &
  \cellcolor[HTML]{DDE7F9}0.17 &
  \cellcolor[HTML]{B9D4AD}0.71 &
  \cellcolor[HTML]{F0BB84}0.76  \\
 &
   &
  E &
  3.07 &
  \cellcolor[HTML]{5D8DE1}0.76 &
  \cellcolor[HTML]{90BB7D}0.81 &
  \cellcolor[HTML]{E99D4D}0.92 &
  2.69 &
  \cellcolor[HTML]{ABC4F0}0.40 &
  \cellcolor[HTML]{D2E4CA}0.65 &
  \cellcolor[HTML]{EFB67A}0.79 &
  3.05 &
  \cellcolor[HTML]{96B5EC}0.50 &
  \cellcolor[HTML]{CEE1C5}0.66 &
  \cellcolor[HTML]{EDB06F}0.82 \\
 &
   &
  A &
  3.50 &
  \cellcolor[HTML]{86AAE9}0.57 &
  \cellcolor[HTML]{A0C590}0.77 &
  \cellcolor[HTML]{F2C392}0.72 &
  \textit{\textbf{4.02}} &
  \cellcolor[HTML]{BFD2F3}0.31 &
  \cellcolor[HTML]{EFF5EC}0.58 &
  \cellcolor[HTML]{ECAA65}0.85 &
  \textit{\textbf{4.58}} &
  \cellcolor[HTML]{E8EFFB}0.12 &
  \cellcolor[HTML]{B5D2A8}0.72 &
  \cellcolor[HTML]{EEB476}0.80\\
 &
   &
  N &
  \textit{\textbf{3.70}} &
  \cellcolor[HTML]{EEF3FC}0.09 &
  \cellcolor[HTML]{BDD7B2}0.70 &
  \cellcolor[HTML]{FFFFFF}0.40 &
  3.41 &
  \cellcolor[HTML]{EEF3FC}0.09 &
  \cellcolor[HTML]{FFFFFF}0.54 &
  \cellcolor[HTML]{F5D0AA}0.65 &
  \textit{\textbf{4.28}} &
  \cellcolor[HTML]{EAF0FB}0.11 &
  \cellcolor[HTML]{C6DCBC}0.68 &
  \cellcolor[HTML]{F0BB84}0.76  \\

\cmidrule{3-15}

 &
  \multirow{-6}{*}{Neutral} &
  Total &
  3.58 &
  \cellcolor[HTML]{ACC5F0}0.40 &
  \cellcolor[HTML]{A0C590}0.77 &
  \cellcolor[HTML]{F1C08C}0.74 &
  \textit{\textbf{3.74}} &
  \cellcolor[HTML]{D3E0F7}0.22 &
  \cellcolor[HTML]{D6E6CF}0.64 &
  \cellcolor[HTML]{F0B981}0.77 &
  \textit{\textbf{4.17}} &
  \cellcolor[HTML]{D9E5F8}0.19 &
  \cellcolor[HTML]{BAD5AE}0.71 &
  \cellcolor[HTML]{EEB172}0.81 \\

\cmidrule{2-15}

 &
   &
  O &
  1.75 &
  \cellcolor[HTML]{5789E0}0.79 &
  \cellcolor[HTML]{A9CA9A}0.75 &
  \cellcolor[HTML]{F3CA9F}0.68 &
  \textit{\textbf{3.27}} &
  \cellcolor[HTML]{C9D9F5}0.26 &
  \cellcolor[HTML]{C1D9B7}0.69 &
  \cellcolor[HTML]{FDF4EB}0.46 &
  \textit{\textbf{3.39}} &
  \cellcolor[HTML]{D0DEF7}0.23 &
  \cellcolor[HTML]{C6DCBC}0.68 &
  \cellcolor[HTML]{FDF4EB}0.46 \\
 &
   &
  C &
  \textit{\textbf{2.36}} &
  \cellcolor[HTML]{6F99E4}0.68 &
  \cellcolor[HTML]{ADCD9F}0.74 &
  \cellcolor[HTML]{FCF2E7}0.47 &
  \textit{\textbf{3.14}} &
  \cellcolor[HTML]{BFD2F3}0.31 &
  \cellcolor[HTML]{E3EEDE}0.61 &
  \cellcolor[HTML]{FDF6EE}0.45 &
  \textit{\textbf{3.31}} &
  \cellcolor[HTML]{E1EAFA}0.15 &
  \cellcolor[HTML]{ADCD9F}0.74 &
  \cellcolor[HTML]{F7DBBE}0.59 \\
 &
   &
  E &
  1.41 &
  \cellcolor[HTML]{3973DA}0.93 &
  \cellcolor[HTML]{A9CA9A}0.75 &
  \cellcolor[HTML]{E99D4D}0.92 &
  1.80 &
  \cellcolor[HTML]{467CDD}0.87 &
  \cellcolor[HTML]{CEE1C5}0.66 &
  \cellcolor[HTML]{EBA45B}0.88 &
  2.18 &
  \cellcolor[HTML]{6693E3}0.72 &
  \cellcolor[HTML]{BDD7B2}0.70 &
  \cellcolor[HTML]{EDB06F}0.82 \\
 &
   &
  A &
  1.66 &
  \cellcolor[HTML]{5385DF}0.81 &
  \cellcolor[HTML]{A0C590}0.77 &
  \cellcolor[HTML]{F6D4B1}0.63 &
  \textit{\textbf{2.34}} &
  \cellcolor[HTML]{6F99E4}0.68 &
  \cellcolor[HTML]{DAE9D4}0.63 &
  \cellcolor[HTML]{F3CA9F}0.68 &
  \textit{\textbf{2.71}} &
  \cellcolor[HTML]{9CB9ED}0.47 &
  \cellcolor[HTML]{D6E6CF}0.64 &
  \cellcolor[HTML]{F9E3CC}0.55 \\
 &
   &
  N &
  1.47 &
  \cellcolor[HTML]{4179DC}0.89 &
  \cellcolor[HTML]{ADCD9F}0.74 &
  \cellcolor[HTML]{F1BF8B}0.74 &
  1.87 &
  \cellcolor[HTML]{437BDC}0.88 &
  \cellcolor[HTML]{DAE9D4}0.63 &
  \cellcolor[HTML]{EDAE6C}0.83 &
  \textit{\textbf{2.64}} &
  \cellcolor[HTML]{82A7E8}0.59 &
  \cellcolor[HTML]{F3F8F1}0.57 &
  \cellcolor[HTML]{F1BF8B}0.74  \\

\cmidrule{3-15}

\multirow{-18}{*}{\rotatebox[]{90}{Personas}} &
  \multirow{-6}{*}{Low} &
  Total &
  1.73 &
  \cellcolor[HTML]{5084DF}0.82 &
  \cellcolor[HTML]{A9CA9A}0.75 &
  \cellcolor[HTML]{F3C99D}0.69 &
  \textit{\textbf{2.48}} &
  \cellcolor[HTML]{80A6E8}0.60 &
  \cellcolor[HTML]{D4E5CD}0.64 &
  \cellcolor[HTML]{F4CEA6}0.66 &
  \textit{\textbf{2.85}} &
  \cellcolor[HTML]{A4BFEE}0.43 &
  \cellcolor[HTML]{CBE0C3}0.67 &
  \cellcolor[HTML]{F5D3B0}0.63  \\

  \midrule
    \multicolumn{3}{c|}{Total} & 
    & 0.73 &	0.81 &	0.78	&
  & 0.58	&0.68	& 0.77		&
  & 0.52	& 0.73	& 0.79 \\
  
\bottomrule

    \multicolumn{15}{l}{* $s_{target}$: one of 3 parts into which the range of [1,5] is divided; [1: low-level persona $\ldots$ 3: neutral-level persona $\ldots$ 5: high-level persona]}
\end{tabular}%
}
\caption{The detailed results of GPT-4o assigned 15 personas. The row of `\textit{High-Level}--\textit{O-Dimension}' is the result of GPT-4o assigned with a persona of \textit{open-minded person}. MEAN indicates averaged characteristic scores, and bold scores in MEAN mean those in ranges of \textbf{\textit{persona-misaligned}} scores. For \accatom, \rcatom, and \icatom, we represent high scores with the darker region.}
\label{tab:main_result_gpt4o}
\vspace{-0.1in}
\end{table*}

\subsection{Comparison with Previous Metrics}

\paragraph{Our metrics capture subtle misalignment and inconsistency not previously covered.}
%These atomic counterparts of ACC and RC exhibit high correlation (Table~\ref{tab:correlation_metrics_old_ours}, r = [0.91, 0.98]), indicating they measure similar underlying aspects. 
\accatom\ and \rcatom\ exhibit high correlation with their counterparts ($r$ = [0.91, 0.98] in \autoref{tab:correlation_metrics_old_ours}), indicating that they measure similar underlying aspects. However, \autoref{fig:comparison_prev_ours} reveals key differences. Some models with low-level personas achieve high ACC but low \accatom, primarily due to persona-misaligned sentences within generated texts. While these texts may score correctly on average, they often contain sentences contradicting the assigned persona. Previous metrics, such as ACC, fail to detect these ``glitches'', leading to inflated scores. Conversely, \accatom\ identifies these deviations, offering a more granular assessment of persona fidelity in long-form generations.

% \paragraph{\icatom\ captures intra-generation consistency not previously covered.} %unique aspects of persona fidelity in generation tasks, which are overlooked by traditional metrics.}
The necessity of \icatom\ as a distinct metric is evident from their low correlations with previous metrics ($r$ = [0.40, 0.37]; refer to Table~\ref{tab:correlation_metrics_old_ours}). It highlights the need for a multi-dimensional and comprehensive approach to persona evaluation, where accuracy, consistency across responses, and internal coherence are assessed together.

% 2. 예시 기반 ACC, AIC, ATRT 분석

\paragraph{Understanding through case study}
Our metrics address limitations in previous evaluation methods by capturing deeper layers of persona fidelity within generated texts.
In the case of \autoref{tab:method_example}, the previous method assesses that both generations are accurate and aligned to the given persona and the model successfully reproduces the generations of the same personality traits.
However, our evaluation framework reveals distinct points: Gen B exhibits low fidelity with \accatom = 0 and \icatom = 0.06, whereas Gen A achieves full scores on both metrics.
%This indicates that Gen B exhibits low persona accuracy and internal consistency across atomic-level segments, which our metrics detect.
Unlike previous evaluation, \accatom\ captures the persona misalignment and \icatom\ identifies fluctuating behavior and low internal coherence in Gen B.
Moreover, the model’s \rcatom\ score becomes 0, highlighting its moderate ability to consistently reproduce atomic-level persona fidelity.
%This case study illustrates how AACC, ATRT, and AIC allow for a more comprehensive evaluation of generative outputs, uncovering distinctions in persona fidelity and internal consistency that remain invisible under existing metrics. By leveraging our advanced metrics, we can achieve a deeper understanding of the stylistic and persona alignment tendencies of generative models.
This case study demonstrates that our metrics---\accatom, \rcatom, and \icatom---enable a more granular evaluation, uncovering distinctions in persona fidelity and internal consistency that traditional metrics fail to capture.

% By incorporating these advanced metrics, we expand the scope of persona fidelity evaluation, offering a more comprehensive framework to analyze the alignment and consistency of persona-assigned models.

\subsection{Understanding Model's Behavior with Diverse Conditions}
\label{sec:results-understanding}
%%%%%
% Referring to the results of GPT-4o presented in Table~\ref{tab:main_result_gpt4o}, we demonstrate how our metrics facilitate a detailed and comprehensive analysis of task-wise and persona-level trends.

In this section, we use GPT-4o as a case study (\autoref{tab:main_result_gpt4o}) to illustrate the utility of \accatom\ and \rcatom. The following analysis demonstrates how these metrics reveal detailed insights into model performance across different tasks and personas.

% Task-Wise Performance
%The results highlight notable performance differences across the Questionnaire, Essay, and Social Media Post tasks. For most personas, GPT-4o achieved higher scores across all metrics in the Questionnaire task compared to the Essay and Social Media Post tasks. 
%For example, the total AACC score for the Questionnaire task was 0.98, compared to 0.93 for Essay and 0.81 for Social Media Post. Similarly, ATRT and AIC scores also decreased slightly for the latter two tasks. The most significant performance drops were observed in the neutral persona group, with AACC scores as low as 0.01 in the Essay task for O$_{neutral}$ and N$_{neutral}$.

% \paragraph{Persona-assigned generations exhibit high accuracy and strong internal consistency in tasks with clear structural and highly contextual cues.}
\paragraph{Persona accuracy and consistency are high in high-contextual, structured tasks.}
The results highlight notable performance differences across the Interview with questionnaires (Questionnaire), Essay, and Social Media Post tasks. GPT-4o generally achieves better \accatom\ and \icatom\ scores in the Questionnaire task compared to the Essay and Social Media Post tasks (\accatom = [0.73, 0.58, 0.52], \icatom = [0.81, 0.68, 0.73], in the order of the mentioned tasks). 
% For instance, the total \accatom\ score for the Questionnaire task is 0.73, compared to 0.58 and 0.52 for the Essay and the Social Media Post tasks, respectively. 
% Similarly, \icatom\ scores follow this trend, reflecting stronger internal consistency within persona-aligned outputs in structured tasks like questionnaires.
% These trends suggest that GPT-4o performs best in tasks with clear structural cues, such as questionnaires, where the task provides explicit guidance.
The Essay task provides scenarios that are indirectly related to personality, and the Social Media Post task offers possible topics for generation. In contrast, the Questionnaire task provides structured questions that explicitly include vocabulary and expressions directly linked to the assigned persona. 
Considering the differences across tasks, the results suggest that persona alignment improves when the model is exposed to contextual cues relevant to the assigned persona.

However, \rcatom\ scores display varied patterns across different personas, and no clear task-dependent trend is observed. While some personas maintain consistent \rcatom\ scores across tasks, others show significant variability. For example, for the \textit{Low E} persona, the model achieves the highest \rcatom\ score in the Questionnaire task, while for the \textit{High A}, it performs best in the Social Media Post task. On the other hand, \textit{Neutral O} exhibited consistent \rcatom\ scores across all tasks.

%These results suggest that GPT-4o performs best in tasks with explicit structural and contextual cues, such as questionnaires, where persona alignment is easier to enforce. On the other hand, ATRT’s varied patterns indicate that the reproducibility of persona fidelity is less influenced by task structure and more by the inherent alignment challenges of each persona.

%subsection 3
\begin{table*}[t!]
    \centering
    \begin{minipage}[t]{0.49\textwidth}
        \centering
        \resizebox{\linewidth}{!}{
        
%\resizebox{\columnwidth}{!}{%
\begin{tabular}{l|rrr}
\toprule
 & \multicolumn{1}{c}{Questionnaire} & \multicolumn{1}{c}{Essay} & \multicolumn{1}{c}{Social Media Post} \\
 Models &
  \multicolumn{1}{r}{\# (valid\%)}  &
  \multicolumn{1}{r}{\# (valid\%)}  &
  \multicolumn{1}{r}{\# (valid\%)}  \\

\hline

\rowcolor[HTML]{D9D9D9} 
Davinci-002                        & 6.4 (52.7) & 10.2 (35.0) & 10.5 (38.4) \\
GPT-3.5-turbo                & 4.6 (98.1) & 12.3 (86.6) &15.7 (90.8) \\
GPT-4o              & 5.5 (95.4) & 14.8 (84.2) & 23.7 (87.8) \\

\hline

\rowcolor[HTML]{D9D9D9} 
Llama-3-8B                    & 8.2 (60.8) & 14.3 (50.1) & 14.3 (55.0) \\
Llama-3-8B-Instruct           & 7.0 (95.2) & 16.7 (67.7) & 24.3 (80.6) \\

\hline

\rowcolor[HTML]{D9D9D9} 
Mistral-7B          & 8.9 (58.2) & 13.3 (51.4) & 15.4 (55.2) \\
Mistral-7B-Instruct & 5.8 (93.4) & 12.4 (88.5) & 21.8 (88.4) \\

\hline

Claude-3-haiku           & 6.6 (84.4) & 17.2 (87.6) & 19.9 (88.0) \\

\bottomrule

\end{tabular}%
%}
% \caption{The average number of atomic sentences in one generation for each task (\#). The numbers in parentheses indicate the proportion of valid sentences among the generated atomic sentences (\%).}
% \label{tab:data_statistics}
% \vspace{-0.2in}

        }
        \caption{The average number of atomic sentences in one generation (\#). The numbers in parentheses indicate the proportion of valid sentences among the atomic sentences (\%).}
        \label{tab:data_statistics}
        % \vspace{-0.2in}
    \end{minipage}
    \hfill
    \begin{minipage}[t]{0.474\linewidth}
        \centering
        \resizebox{\linewidth}{!}{% \begin{table}[t]
% \centering
% \resizebox{\columnwidth}{!}{
\begin{tabular}{lcc|ccc}

\toprule

    & & & \accatom & \icatom & \rcatom \\
Models &
    Inst-FT & RLHF &
  [0,1]&
  [0,1] &
  [-1,1] \\

\hline

\rowcolor[HTML]{D9D9D9} 
Davinci-002 &
  \ding{51}&
   &
  0.39 &
  0.64 &
  0.56  \\
GPT-3.5-turbo                 & \ding{51}      & \ding{51}   & 0.60  & \textbf{0.75}         & 0.79                     \\
GPT-4o                 & \ding{51}      & \ding{51}   & 0.61      & 0.74        & 0.78                     \\

\hline

\rowcolor[HTML]{D9D9D9} 
Llama-3-8B &
   &
   &
  0.41&
  0.60 &
  0.64   \\
Llama-3-8B-Instruct           & \ding{51}      & \ding{51}   & \textbf{0.65}  & 0.70           & \textbf{0.82}                   \\

\hline

\rowcolor[HTML]{D9D9D9} 
Mistral-7B &
   &
   &
  0.41 &
  0.59 &
  0.67 
  \\
Mistral-7B-Instruct & \ding{51}      &      & 0.58    & 0.69          & 0.80                  \\

\hline
Claude-3-haiku            & \ding{51}      & \ding{51}   & 0.59      & 0.71          & 0.69                  \\
\bottomrule

\end{tabular}%
% }
% % \caption{The overall experimental results. For all metrics, the bigger the score, the better the persona fidelity. The scores for the best performances are shown in \textbf{bold}.}
% % \label{tab:main_result}
% % \vspace{-0.2in}
% \end{table}
        }
        \caption{The overall experimental results. Inst-FT means an instruction-tuned model. For all metrics, the bigger the score, the better the persona fidelity. The scores for the best performances are shown in \textbf{bold}.}
        \label{tab:main_result}
        \vspace{-0.2in}
    \end{minipage}
\end{table*}

% Persona-Level Trends
\paragraph{Models demonstrate strong persona fidelity with clearly defined and socially desirable personas.} % RLHF-aligned personas.}
GPT-4o demonstrates strong task-agnostic persona fidelity for the high-level persona group, consistently achieving the highest scores across all tasks. For instance, the model achieves almost perfect \accatom\ scores and high \icatom\ and \rcatom\ values with \textit{High N (emotionally stable)} persona. %, showcasing the model’s ability to align with clearly defined and well-established personas.
However, the model displays the weakest fidelity, particularly in \accatom, for neutral-level personas.
It highlights challenges in maintaining alignment for personas with less distinct or ambiguous characteristics. For example, GPT-4o achieves low \accatom\ scores with \textit{Neutral O} and \textit{Neutral N} (0.14 and 0.09, respectively) in the Questionnaire task, with similarly poor performance in Essay and Social Media Post tasks.

For the low-level personas, the model exhibits moderate but unstable fidelity across tasks, suggesting potential inconsistencies in persona alignment. For instance, while GPT-4o assigned \textit{Low A} persona maintains relatively high \accatom\ scores in the Questionnaire task (0.81), performance dropped significantly in the Social Media Post task (0.47). This performance variation of the low-level personas indicates that they pose greater challenges for the model in maintaining consistent alignment across tasks.
Notably, not only GPT-4o but also most tuned models demonstrate low persona fidelity for personas that are not socially preferred, including \textit{Neutral O} (neither open nor close-minded), \textit{Neutral N} (neither emotionally stable nor neurotic), \textit{Low O} (close-minded), and \textit{Low C} (careless). %These personalities contradict the personality traits or values (e.g., helpfulness, honesty, and harmlessness~\cite{askell2021general, bai2022training}) that the models were aligned with during RLHF training. This suggests that the personas, which differ from the intended alignment goals of RLHF, were unable to override the model's pre-trained alignment behaviors.
Given that RLHF and other alignment techniques rely on human preferences, it is plausible that responses reflecting helpfulness, honesty, and harmlessness~\cite{askell2021general, bai2022training} were consistently favored in the training data. Thus, this trend suggests that models may have been implicitly guided toward socially desirable traits during alignment training. 
%This trend is supported by TC, which shows high stability for the \textit{High} persona group (0.91 on average) but lower consistency for \textit{Neutral} and \textit{Low} groups (0.76 on average for both). This disparity underscores the model's limitations in generalizing persona fidelity for less distinct or more challenging personas.

% Discussion
% These results emphasize the importance of evaluating persona fidelity not only in terms of accuracy but also in consistency across diverse dimensions. Metrics like AICC, ATRT, and TC provide insights into areas where traditional metrics fail, capturing nuances in atomic-level alignment and robustness across tasks. This multi-dimensional approach reveals areas for improvement in handling neutral and low personas, as well as challenges in free-form generative tasks.

\subsection{Model Comparison}

We perform a comprehensive evaluation of the models and report the results in \autoref{tab:main_result}.
%% Atomic Persona Fidelity
The tuned models consistently outperform their base counterparts in all atomic-level persona fidelity scores. This result coincides with their lower valid generation rates (\autoref{tab:data_statistics}), indicating their weaker ability to follow instructions effectively.
% gpt계열은 AIC에서 강함을 보여주었고 (내부 일관적인 생성) -> 혼란스럽지 않게 일정한/일관된 성격으로 하나의 generation을 작성하는 능력이 강함. 더 일관되고 자연스러운 generation을 기대할 수 있음.

GPT models and Claude show strong performance in \icatom\ (0.71 -- 0.75), reflecting their ability to produce internally coherent generations.
% llama 계열은 aacc와 atrt에서 강함을 보여주었다. -> 주어진 성격에 더 정확한 문장을 생성. 여러 번 반복했을 때 비슷한 성격 점수 분포의 generation을 만들어낼 수 있음.
The instruction-tuned LLaMA model achieves the highest \accatom\ and \rcatom\ scores, indicating their precision in generating persona-aligned sentences and maintaining a consistent distribution of atomic-level personality scores across repeated generations.
% On the other hand, the base models exhibited relatively poor persona alignment, which coincided with their lower valid generation rates (see Table~\ref{tab:data_statistics}), indicating their weaker ability to follow instructions effectively.
These observations emphasize the diversity in model strengths, offering a broad range of choices for specific applications.
Our findings highlight the importance of evaluation metrics in assessing persona fidelity at diverse dimensions.

\section{Analysis}

\begin{table}[t]
    \centering
    \small
    \begin{tabular}{l|rrr}
    \toprule 
    & \accatom &  \icatom & \rcatom \\
    \midrule
        \accatom &	1 & 0.45 &	0.51  \\
	 \icatom &	-	& 1	& 0.44 \\
	 \rcatom  &	- &	-	& 1	\\
    \bottomrule
    \end{tabular}
    \caption{Pearson correlation coefficient ($r$) between scores of the proposed metrics.}
    \label{tab:correlation_metrics_ours}
    \vspace{-0.1in}
\end{table}
%\input{Table/6Analysis/case_study}

% \paragraph{Orthogonality of Our Metrics}
% \input{Table/6Analysis/correlation_metrics_ours}

%1.  \accatom != consistency scores we should consider consistency of diverse dimension for persona fidelity because consistency scores do not always follow accuracy performance.
\paragraph{Our proposed metrics capture distinct aspects of persona fidelity.}
% We conduct correlation tests on the scores of our metrics.
% Accuracy-based metrics (\accatom) and consistency-based metrics (\icatom, \rcatom) show moderate correlations ($r_{\text{ACC}_{\text{atom}}, \text{IC}_{\text{atom}}}$ = 0.45 and $r_{\text{ACC}_{\text{atom}}, \text{RC}_{\text{atom}}}$ = 0.51), emphasizing the importance of considering multiple dimensions beyond accuracy.

% This finding underscores the need to evaluate persona fidelity through diverse consistency dimensions, as high persona accuracy does not necessarily imply strong internal coherence or reproducibility.
By conducting correlation tests on our metric scores, we observe that the accuracy-based metric (\accatom) and consistency-based metrics (\icatom\ and \rcatom) exhibit moderate correlations ($r$ = [0.45, 0.51]; \autoref{tab:correlation_metrics_ours}).
It emphasizes the importance of evaluating multiple dimensions beyond accuracy, as high persona accuracy does not necessarily imply strong internal coherence or reproducibility.

Furthermore, it reveals that internal consistency (\icatom) and retest consistency (\rcatom) are not strongly correlated ($r=0.44$). This suggests that maintaining a stable persona within a single response does not guarantee consistency across repeated generations. A model may produce internally coherent outputs while shifting its characteristics across different generations, reinforcing the need to assess both intra-response (\icatom) and inter-response (\rcatom) consistency separately.

%2. aTRT != TC ... 동일한 태스크를 일관적이게 잘 수행하는 능력이랑 across task로 consistent한 task-agnostic robustness는 다른 능력인가보다
% Distinction Between Task-Specific and Task-Agnostic Consistency
% Moreover, the correlation between \rcatom\ and $\sigma_{\text{\accatom}}$ ($r$ = 0.27) suggests that maintaining consistency within a single task (\rcatom) differs from achieving persona fidelity across multiple tasks ($\sigma_{\text{\accatom}}$). This highlights a fundamental distinction between task-specific consistency---the model's ability to generate stable and unchanging responses within a given task---and task-agnostic robustness, which reflects the model’s ability to generalize persona fidelity across diverse generative contexts.

% 3. Implications for Persona Fidelity Evaluation
%By incorporating these consistency-focused metrics, our framework broadens the evaluation perspective, enabling a more holistic analysis of persona fidelity that goes beyond surface-level accuracy. This ensures a deeper understanding of how well models maintain fidelity in both atomic-level and task-level consistency across diverse generative contexts.
These findings reinforce the orthogonality of our proposed metrics and the necessity of evaluating diverse dimensions of persona fidelity.
%, demonstrating that persona fidelity should not be assessed solely based on accuracy but also through reproducibility, internal coherence, and task consistency. By integrating these consistency-focused metrics, our framework provides a more comprehensive and multi-faceted evaluation of persona fidelity, ensuring a deeper understanding of how well models maintain persona alignment across atomic and task-level consistency measures.
By integrating these metrics, our evaluation framework provides a more comprehensive and multi-faceted evaluation of persona fidelity, ensuring a deeper understanding of how well models maintain persona alignment across atomic units. % and task-level consistency measures.

\paragraph{The length of generations has a minimal impact on persona fidelity.}
% 원래 제너레이션 길이가 길어지면 내부 일관성에 문제가 있다는 / generation의 뒤에 오는 문장이 처음과 연관이 없어진다는 (coherence 유지 어려움) 선행연구 있음.
% 우리 aic 점수도 generation 길이에 의존하여 결정되는 값일까?
% 우리는 generated sentences의 개수와 AIC 점수의 correlation을 보았다.
% 확인해보니 AIC 점수는 generation의 수, 유효한 generation의 수 모두와 매우 낮은 상관을 보였다 (r=[-.31, -0.12]).
% 뿐만 아니라 생성 길이는 AACC, ATRT와도 낮은 상관을 보였다 (r=[0.20, -0.12]).
% 우리의 태스크처럼 100-300 개의 단어로 long generation을 생성하는 과제에서는 길이가 길어짐에 따라 persona assigned model의 fidelity가 크게 하락하거나 generation performance의 tendency 변화하는 것으로 나타나지 않았다.
Previous studies have suggested that as the length of generated text increases, internal consistency issues arise, resulting in reduced coherence between the earlier and later parts of the text~\cite{sun-etal-2021-long, deng2022model, krishna-etal-2022-rankgen}. This issue makes it challenging for models to maintain persona fidelity over extended generations.
To investigate whether our \icatom\ score is influenced by generation length, we analyze its correlation with the number of generated atomic sentences. The results indicate that \icatom\ scores have very low correlations with both the number of generated sentences and the number of valid sentences ($r$ = [-0.31, -0.12]). 
Additionally, generation length shows weak correlations with \accatom\ and \rcatom\ ($r$ = [0.20, -0.12]), indicating that the fidelity of persona-assigned models does not degrade as text length increases. In tasks requiring 100-300 word-long responses, we do not observe a strong tendency for longer generations to result in a substantial decline in persona fidelity or overall generation performance.

\section{Conclusions}

In this study, we highlight the challenge of subtle Out-of-Character (OOC) behavior in persona-assigned large language models (LLMs) during open-ended text generation. To address this, we introduce an atomic-level evaluation framework to assess persona fidelity in LLMs. Our approach captures subtle inconsistencies at the sentence level that previous response-level metrics overlook, offering a finer-grained understanding of persona alignment and enabling a more precise measurement of persona fidelity. Through experiments, we demonstrate that our framework effectively identifies Out-of-Character behaviors and provides a deeper understanding of how well LLMs maintain their assigned personas throughout the generated text. Our approach successfully detects these misalignments, highlighting challenges in maintaining consistent persona expression across diverse conditions. 
We hope that our framework serves as a foundation for future work in diagnosing, benchmarking, and ultimately improving persona consistency in personalized language generation.

\section*{Ethical Considerations}

% persona assigning / role playing 기술이 악용되어 사용되어서는 안 된다.

% persona fidelity를 더욱 발전시키는 연구가 중요하지만 모델의 궁극적인 윤리적 목적을 앞서면 안 될 것. persona fidelity와 더불어서 사용자 자신이나 공동체를 위험에 빠뜨리게 하는 사용자의 의도를 잘 구분해내는 기술의 발전 필요성을 우리는 주장한다.
LLMs, while demonstrating impressive capabilities in generating human-like content, raise critical ethical considerations due to their potential for misuse. While personified agents such as persona-assigned models and role-playing agents can offer more engaging daily interactions, this potential also necessitates careful consideration of the risks to individuals, communities, and society. Researchers and developers should be mindful of this potential and strive to uphold AI ethics standards. 

% Also, it is crucial to emphasize that this work is primarily a scientific inquiry into LLM expressivity and human personality perception through LLM generations.  
% In addition to enhancing persona fidelity, we urge the development of techniques that can accurately identify and prevent user intentions that may endanger the users themselves or the community.

It is important to emphasize that this study is primarily a scientific investigation into LLM expressivity and human personality perception through LLM-generated responses. While persona-assigned model responses may be evaluated as faithful, they can also be perceived as biased. Since our study focuses on personality traits, it is not directly entangled with more sensitive biases, such as those related to demographics or social characteristics. However, in applications involving personas tied to social identities, careful interpretation of results is essential to avoid reinforcing unintended biases.  
Beyond simply improving persona fidelity, we advocate for the development of techniques that can accurately identify and prevent harmful user intentions that may endanger individuals or communities. Ensuring responsible AI deployment requires balancing faithful persona alignment with safeguards against misuse, particularly in scenarios where personas can influence user interactions in unintended ways.

\section*{Limitations}

% This study has several limitations.
In this section, we discuss the limitations of our study and outline potential directions for future research.
First, our study focuses solely on the personality domain for both persona creation and evaluation. Personality was chosen as the primary domain due to its well-established trait structure and its prominence in prior persona-related research, which allowed for controlled and consistent analysis. However, our evaluation framework is inherently domain-agnostic and can be applied to any persona dimensions with quantifiable characteristics. Future work could extend this framework to other domains such as social values or political leanings to gain a deeper understanding of persona LLMs.

Second, the reliance on LLMs for both scoring and filtering introduces potential biases and inaccuracies. Results may vary depending on the specific scoring LLM used, and any inherent biases or errors in LLM could influence the final evaluations. 
To address this concern, we conducted a human evaluation comparing LLM-based scores with human judgments and observed strong alignment, supporting the validity of our framework.
Nonetheless, future research could consider human-in-the-loop setups or ensemble scoring methods that combine multiple LLM evaluators to enhance robustness.

Third, while atomic-level analysis enables a fine-grained evaluation, it may not fully capture the full context of an entire response. Our metrics are designed to complement, not replace, response-level evaluation by revealing subtle inconsistencies. In addition, since our experiments adopt sentence-level segmentation as the atomic unit, intra-sentence inconsistencies may remain undetected. We emphasize that this segmentation level is an experimental design choice, not a fixed constraint of the framework. Future work could explore alternative segmentation strategies---such as phrase-level, information-level, or discourse-level units---to more effectively capture nuanced inconsistencies.

Fourth, one potential concern is whether individual sentences sufficiently reflect persona characteristics. Although not all sentences in real-world dialogue strongly reflect persona traits, our experimental design was intentionally constructed to elicit persona-relevant generation through carefully designed tasks, as described in Section~\ref{section:experimental_setup-tasks}. As shown in \autoref{tab:data_statistics}, over 80\% of sentences in our generation contain identifiable personality signals, supporting the appropriateness of sentence-level evaluation within our setup.

Lastly, while our study does not propose direct mitigation strategies, our evaluation framework offers actionable insights that can inform future work aimed at improving persona consistency in LLMs.

\section*{Acknowledgments}
This work was supported by Institute of Information \& communications Technology Planning \& Evaluation (IITP) grant funded by the Korea government (MSIT) (No.RS-2022-II220184, Development and Study of AI Technologies to Inexpensively Conform to Evolving Policy on Ethics).
We used AI assistants, including ChatGPT\footnote{\url{https://chatgpt.com/}} and Grammarly\footnote{\url{https://www.grammarly.com}}, to support the writing and coding processes.

% Bibliography entries for the entire Anthology, followed by custom entries
\normalem % delete underline 
%\bibliography{anthology,custom}
% Custom bibliography entries only
\bibliography{custom}

\clearpage
\appendix
\section{Appendix for Experimental Setup}
\label{sec:appendix_setup}

The code and prompts used in our experiments are publicly available on our GitHub repository\footnote{\url{https://github.com/ddindidu/atomic-persona-evaluation}}.

\subsection{Tasks}
\label{appendix:setup-tasks}

\begin{table}[thb!]
    \centering
    \caption{Instruction for the Interview with Questionnaire task. Dim. represents the dimension of personality.}
    \label{tab:task_questionnaire}

    \resizebox{\linewidth}{!}{
    \small
    \begin{tabular}{l}
        \toprule
        \multicolumn{1}{c}{Task prompt~\cite{wang-etal-2024-incharacter}} \\
        \midrule
        Describe your personality under 100 words. \\
        \{\textit{question}\} \\
        Response: \\
        \bottomrule
    \end{tabular}
    }    

    \small
    \resizebox{\linewidth}{!}{
    \begin{tabular}{cl}
        \toprule
        Dim. & \textit{question}~\cite{trull1998structured} \\
        \midrule
            & Do you have a rich vocabulary?\\
            & Do you have difficulty understanding abstract ideas?\\
            & Do you have a vivid imagination?\\
            & Do you think you are not interested in abstract ideas?\\
         O   & Do you have excellent ideas?\\
            & Do you think you do not have a good imagination?\\
            & Are you quick to understand things?\\
            & Do you use difficult words?\\
            & Do you spend time reflecting on things?\\
            & Are you full of ideas?\\
        \midrule
         &Are you always prepared?\\
            &Do you leave your belongings around?\\
            &Do you pay attention to details?\\
            &Do you make a mess of things?\\
            &Do you get chores done right away?\\
        C    &Do you often forget to put things back in their proper place?\\
            &Do you like order?\\
            &Do you shirk your duties?\\
            &Do you follow a schedule?\\
            &Are you exacting in your work? \\
        \midrule
        &Are you the life of the party?\\
            &Do you think you don't talk a lot?\\
            &Do you feel comfortable around people?\\
            &Do you keep in the background?\\
        E    &Do you start conversations?\\
            &Do you have little to say?\\
            &Do you talk to a lot of different people at parties?\\
            &Do you think you don't like to draw attention to yourself?\\
            &Do you think you don't mind being the center of attention?\\
            &Are you quiet around strangers?\\
        \midrule
         &Do you feel little concern for others?\\
            &Are you interested in people?\\
            &Do you insult people?\\
            &Do you sympathize with others' feelings?\\
            &Do you think you are not interested in other people's problems?\\
        A    &Do you have a soft heart?\\
            &Do you think you are not really interested in others?\\
            &Do you take time out for others?\\
            &Do you feel others' emotions?\\
            &Do you make people feel at ease?\\
        \midrule
         &Do you get stressed out easily?\\
            &Are you relaxed most of the time?\\
            &Do you worry about things?\\
            &Do you seldom feel blue?\\
            &Are you easily disturbed?\\
        N    &Do you get upset easily?\\
            &Do you change your mood a lot?\\
            &Do you have frequent mood swings?\\
            &Do you get irritated easily?\\
            &Do you often feel blue?\\
        \bottomrule
    \end{tabular}
    }

\end{table}

\begin{table}[thb!]
    \centering
    \caption{Instructions for the Essay task. Dim. represents the dimension of personality.}
    \label{tab:task_essay}

    \small
    
    \resizebox{\linewidth}{!}{
    \begin{tabular}{p{\linewidth}}
    %\\
    \midrule
    \multicolumn{1}{c}{Task prompt~\cite{kwantes2016assessing}}    \\
    \midrule
    Please imagine that you are in the following scenario. Knowing yourself as you do, describe how you would feel and what you would do (not what you should do) in each situation. Do not spend too much time thinking about your answer–instead, start writing, and express what comes to mind. Do not labour over your spelling and grammar. Your writing will not be graded. Instead, it is important that you express your thoughts while you have them. Please try to write between 200 and 300 words.\\
    \{\textit{context}\}\\
    Response:\\
    \bottomrule
    \end{tabular}
    }

    \resizebox{\linewidth}{!}{
    \begin{tabular}{cp{0.9\linewidth}}
    \toprule
    Dim. & \textit{context}~\cite{kwantes2016assessing}\\
    \midrule
    O & You have won an Air Canada paid vacation package for one person to any destination in the world. Your package includes round trip plane tickets, accommodations for any type of lodging, and \$5000 spending money. Assuming that you were available to go, where would you choose to go and why? \\
    \midrule
    C & You're working alone late at the office and you notice a strange smell and a hazy mist hanging in the air of the corridor. You suspect it's some gas or vapor leak from some equipment or machinery in the building. You have no idea whether the leaked vapor is hazardous. As honestly as possible, describe what you would do in this situation. \\
    \midrule
    E & Your friend wants you to attend an important party to which he/she has been invited. You have never met the host, and are not very familiar with the crowd of people who will be attending the party, but you agree to meet your friend at the party at 9:00 pm anyway. When you arrive there, you realize that your friend is late. How would you feel, and what would you do while you waited for your friend? \\
    \midrule
    A & Your housemate decides to paint her bedroom a new colour. One night, when you come home from class, you discover that she also painted your room in the same colour because she had paint left over and didn't want it to go to waste. As realistically as possible, describe how you would feel and how you would you handle the situation. \\
    \midrule
    N & You have developed an email friendship with someone. In your latest email, you ask your friend a more personal question. Your friend usually replies quite promptly, but has taken unusually long to reply to your latest questions. Discuss how you would interpret this long period of silence, how you would react and what you would do about it? \\
    \bottomrule

    \end{tabular}
    }
    
\end{table}

\begin{table}[thb!]
    \centering
    \caption{Instruction for Social Media Post task.}
    \label{tab:task-socialmediapost}

    \small
    
    \resizebox{\linewidth}{!}{
    \begin{tabular}{p{\linewidth}}
    \toprule
    \multicolumn{1}{c}{Task prompt~\cite{safdari2023personality}}    \\
    \midrule
    Generate a Facebook status update. The update must be verbose and reflect your character. The updates should cover, but should not be limited to, the following topics: work, family, friends, free time, romantic life, TV / music / media consumption, and communication with others.\\
    Response:\\
    \bottomrule
    \end{tabular}
    }

\end{table}
As mentioned in Section~\ref{section:experimental_setup}, the Questionnaire task and the Essay task provide different inputs (a different question set for the Questionnaire task and a different context for the Essay task) to a model depending on the assigned persona of the model~\cite{wang-etal-2024-incharacter, jiang2024evaluating}.

Unlike other tasks, the Questionnaire task presents all 10 questions in a single run. In this task, \accatom\ and \icatom\ are obtained as follows: 1) computing atomic-level accuracy and internal consistency for each of the 10 responses individually and 2) averaging them to get the \accatom\ and \icatom\ for the one run. For \rcatom, we aggregate the score distributions of all 10 responses within a single run to form a merged score distribution. Then, \rcatom\ is calculated based on the Earth Mover’s Distance (EMD) between score distributions across multiple runs.

We provide detailed instructions for the three generation tasks in Table~\ref{tab:task_questionnaire}, \ref{tab:task_essay}, and \ref{tab:task-socialmediapost}.

\subsection{Persona Assignment}
\label{appendix:setup-persona}

\begin{table}[!t]
    \centering
    \caption{Instructions for persona assignment. \{\textit{persona}\} can be replaced with any persona entity from the 15 personas below.}
    \label{tab:persona_instruction}
    \resizebox{\linewidth}{!}{
    \begin{tabular}{p{\linewidth}}
        \toprule
        \multicolumn{1}{c}{Persona Assigning Instruction}\\
        \midrule
        $\bullet$ You are a \{\textit{persona}\}.\\
        $\bullet$ You are \{\textit{persona}\}. Your responses should closely mirror the knowledge and abilities of this persona.\\
        $\bullet$ Adopt the identity of \{\textit{persona}\}. Answer the questions while staying in strict accordance with the nature of this identity.\\
        $\bullet$ Take the role of \{\textit{persona}\}. It is critical that you answer the questions while staying true to the characteristics and attributes of this role.\\
        $\bullet$ Speak exactly like \{\textit{persona}\}. Your answer should copy the style of {0}, both the writing style and words you use.\\
        $\bullet$ Speak like \{\textit{persona}\}.\\
        \bottomrule 
    \end{tabular}
    }

    \resizebox{\linewidth}{!}{
    \small
    \begin{tabular}{ccp{0.6\linewidth}}
    \toprule
        Levels & Dim. & \textit{persona}\\
    \midrule
         & O & open person\\
         & C & conscientious person \\
    High     & E & extroverted person\\
         & A & agreeable person\\
         & N & emotionally stable person\\
    \midrule
         & O & neither open nor close-minded person\\
         & C & neither conscientious nor careless person \\
    Neutral     & E & neither extroverted nor introverted person \\
         & A & neither agreeable nor disagreeable person \\
         & N & neither emotionally stable nor neurotic person\\
    \midrule
         & O & close-minded person\\
         & C & careless person \\
    Low     & E & introverted person\\
         & A & disagreeable person\\
         & N & neurotic person\\
    \bottomrule
    \end{tabular}
    }
    
\end{table}
%\input{Table/Appendix_Experiments/personas}
%1. persona list
Our fifteen personality persona entities are in Table~\ref{tab:persona_instruction}.
Our personality personas were designed based on previous studies~\cite{jiang2024evaluating, huang-etal-2024-reliability}.
%2. promt for task
We utilize six persona-assigning prompts referenced in previous studies~\cite{cheng2023marked, deshpande-etal-2023-toxicity, wan2023personalized, gupta2024bias, ko2024different, salewski2024context, shin-etal-2024-ask, shu-etal-2024-dont}.
We prompt a persona-assigning instruction as the \texttt{system} prompt of the model, and each of the five iterations employs a different prompt.

\subsection{Models and Hyperparameters}
\label{appendix:setup-model}
To reproduce our experimental results, we provide our experimental setups for the models.
We employ the four model families, including 12 LLMs: GPTs, LLaMAs, Mistrals, and Claudes.

\paragraph{GPT family}
\begin{itemize}
    \item \textbf{Base}: davinci-002 (utilized via OpenAI\footnote{\href{https://platform.openai.com/}{https://platform.openai.com/}}. Default hyperparameter settings of completion function; temperature=1, top\_p=1, presence\_penalty=0, frequency\_penalty=0, max\_tokens=100, stop=null)
    \item \textbf{Tuned}: gpt-3.5-turbo-0125, gpt-4-turbo-2024-04-09, gpt-4o-2024-08-06, and gpt-4o-mini-2024-07-18 (utilized via OpenAI. OpenAI's chat completion function's default settings: temperature=1, top\_p=1, presence\_penalty=0, frequency\_penalty=0, stop=null)
\end{itemize}

We set the temperature to the default setting to investigate the models' behavior, which is commonly used in general user interactions.

\paragraph{LLaMA family}
\begin{itemize}
    \item \textbf{Base}: meta-llama/Llama-3-8b\footnote{\href{https://huggingface.co/meta-llama/Meta-Llama-3-8B}{https://huggingface.co/meta-llama/Meta-Llama-3-8B}}: utilized via vLLM\footnote{\href{https://docs.vllm.ai/en/latest/}{https://docs.vllm.ai/en/latest/}} (temperature=[0.85,1], max\_tokens=300).
    \item \textbf{Tuned}: meta-llama/Meta-Llama-3-8B-Instruct and meta-llama/Meta-Llama-3-70B-Instruct, via DeepInfra\footnote{\href{https://deepinfra.com/}{https://deepinfra.com/}} (max\_tokens=512, temperature=1, top\_p=1, top\_k=0, repetition\_penalty=1, presence\_penalty=0, frequency\_penalty=0)
\end{itemize}
% We use two LLaMA models, meta-llama/Meta-Llama-3-8B-Instruct and meta-llama/Meta-Llama-3-70B-Instruct, via DeepInfra\footnote{\href{https://deepinfra.com/}{https://deepinfra.com/}}.
% We set the hyperparameters as follows: max\_tokens=512, temperature=1, top\_p=1, top\_k=0, repetition\_penalty=1, presence\_penalty=0, frequency\_penalty=0.

\paragraph{Mistral family}
\begin{itemize}
    \item \textbf{Base}: mistralai/Mistral-7B-v0.3\footnote{\href{https://huggingface.co/mistralai/Mistral-7B-v0.3}{https://huggingface.co/mistralai/Mistral-7B-v0.3}}: utilized via vLLM.
    \item \textbf{Tuned}: Mistral-7B-Instruct-v0.3 served by DeepInfra, applying the same hyperparameters as those used for the LLaMa models.
\end{itemize}
%We use mistralai/Mistral-7B-Instruct-v0.3 served by DeepInfra, applying the same hyperparameters as those used for the LLaMa models.
\paragraph{Claude family}
\begin{itemize}
    \item \textbf{Tuned}: claude-3-haiku-20240307 and claude-3-sonnet-20240229 from Anthropic\footnote{\href{https://docs.anthropic.com/}{https://docs.anthropic.com/}}, with max\_tokens set to 1024 and temperature set to 1.
\end{itemize}
% We employ claude-3-haiku-20240307 and claude-3-sonnet-20240229 from Anthropic\footnote{\href{https://docs.anthropic.com/}{https://docs.anthropic.com/}}, with max\_tokens set to 1024 and temperature set to 1.

For all models, we prompt a persona instruction in the \texttt{system} prompt and a task instruction in the \texttt{user} prompt.

\subsection{LLM-based Scoring and Filtering}
\label{appendix:setup-scoring model}

\begin{table}[t!]
    \centering
    \caption{Instructions for scoring and filtering. \textit{atomic unit} is $a_i$ divided from a long text. `\textit{9)}' in \textit{options} is an option for filtering.}
    \label{tab:scoring_instruction}
    
    \resizebox{\linewidth}{!}{
    \small
    \begin{tabular}{lp{0.8\linewidth}}
    \toprule
    Task & \multicolumn{1}{c}{Scoring instruction}    \\
    \midrule
    Questionnaire &
        What would be the \{\textit{dimension}\} score for the following response to the question?\newline
        Question: \{\textit{question}\}\newline
        Response: \{\textit{atomic unit}\}\newline
        Options: \{\textit{options}\}\newline
        Reply with a single number:
    \\
    \midrule
    %\\
    %\hline
    %\multicolumn{1}{c}{Scoring instruction for Essay task}    \\
    %\hline
    Essay & 
        Rate the personality of the writer of the following essay: ``\{\textit{atomic unit}\}''.\newline
        \{\textit{options}\}\newline
        Reply with a single number:
    \\
    \midrule
    % \\
    % \hline
    % \multicolumn{1}{c}{Scoring instruction for SNS task}    \\
    % \hline
    Social Media &
    Rate the personality of the writer of the following social media post: ``\{\textit{atomic unit}\}''.\newline
    \{\textit{options}\}\newline
    Reply with a single number:
    \\
    \bottomrule
    \end{tabular}
    }

    \resizebox{\linewidth}{!}{
    \small
    \begin{tabular}{clp{0.65\linewidth}}
    \\
    \toprule
        Dim. & \textit{dimension} &  \textit{options} \\
    \midrule
        O & openness & 1) Very close-minded. 2) Moderately close-minded. 3) Neither open-minded nor close-minded. 4) Moderately open-minded. 5) Very open-minded. 9) None of the above.\\
    \midrule
        C & conscientiousness & 1) Very careless. 2) Moderately careless. 3) Neither conscientious nor careless. 4) Moderately conscientious. 5) Very conscientious. 9) None of the above.\\
    \midrule
        E & extroversion & 1) Very introverted. 2) Moderately introverted. 3) Neither extroverted or introverted. 4) Moderately extroverted. 5) Very extroverted. 9) None of the above.\\
    \midrule
        A & agreeableness & 1) Very disagreeable. 2) Moderately disagreeable. 3) Neither agreeable nor disagreeable. 4) Moderately agreeable. 5) Very agreeable. 9) None of the above.\\
    \midrule
        N & neuroticism & 1) Very neurotic. 2) Moderately neurotic. 3) Neither neurotic nor emotionally stable. 4) Moderately emotionally stable. 5) Very emotionally stable. 9) None of the above.\\
    \bottomrule
    \end{tabular}
    }
\end{table}

\begin{table*}[tb!]
    \centering
    \caption{Example of the human evaluation guideline and sentence pairs. Annotators were given a description of the target personality trait (extroversion in this example) and asked to choose which of the two atomic sentences better expressed a higher level of the target trait.}
    \label{tab:human_eval_example}
    \small
    \begin{tabular}{ccc}
    \toprule
    \multicolumn{3}{c}{\textbf{Description}} \\
    \midrule
    \multicolumn{3}{p{0.95\linewidth}}{People with an extroverted personality are individuals who passionately seek out others and prefer to interact with their environment. Extroversion has aspects such as friendliness, sociability, confidence, active energy, stimulation/excitement seeking, and cheerfulness. The given sentences are responses to interview questions that describe your personality. Compare the two sentences and choose which sentence shows more extroverted characteristics.} \\
    \toprule
    \textbf{Text 1}    & \textbf{Text 2}  & \multicolumn{1}{p{0.2\linewidth}}{Which one do you think is more \textbf{extroverted}?} \\
    
    \midrule
    
    \multicolumn{1}{p{0.35\linewidth}}{I'm more of a reflective and introspective person, often finding energy and inspiration in solitude or small, meaningful interactions.}     & 
    \multicolumn{1}{p{0.35\linewidth}}{I thrive on social interactions and enjoy connecting with people from all walks of life.} & 
    \{\texttt{1 or 2}\} \\
    
    \midrule
    
    \multicolumn{1}{p{0.35\linewidth}}{I thrive on social interactions and enjoy connecting with people from all walks of life.}     & 
    \multicolumn{1}{p{0.35\linewidth}}{I'm more of a reflective and introspective person, often finding energy and inspiration in solitude or small, meaningful interactions.} & 
    \{\texttt{1 or 2}\} \\
    
    \bottomrule
    \end{tabular}
\end{table*}
To measure persona fidelity with atomic-level evaluation, we divide a generation into sentences using \texttt{nltk.tokenize.sent\_tokenize}\footnote{\href{https://www.nltk.org/}{https://www.nltk.org/}}.
For the scoring model, we utilize GPT-4o-2024-08-06\footnote{Accessed via OpenAI API} with temperature = 0 to ensure deterministic evaluation. This setting allows for consistent and reproducible results, minimizing variance in the scoring process.
We conducted scoring and filtering processes using the same scoring model via prompting. If a sentence lacked sufficient personality-related information, it was filtered out by assigning a score of 9 (filtering option).
We design our scoring instruction motivated by \citet{wang-etal-2024-incharacter}, and the detailed prompts are in Table~\ref{tab:scoring_instruction}.

\subsection{Human Validation}
\label{appendix:setup-human validation}

The human evaluation aims to assess the extent to which human annotators agree with the scores assigned by the LLM evaluator. For each of the five personality traits, we select 50 sentence pairs, resulting in 250 pairs in total. Six annotators with native-level English proficiency are recruited and asked to rank which of the two sentences in a pair more strongly expresses the target personality trait (refer to \autoref{tab:human_eval_example} for details). To measure alignment, we compute the rank correlation between the human-assigned ranks and the LLM-based scores using Kendall’s $\tau_A$~\cite{kendall1938new}.

\section{Appendix for Experimental Results}
\label{sec:appendix_results}

\subsection{Experimental Results}
\label{sec:appendix_results...experimental_results}

% overall

\begin{table}[t]
\caption{The average number of atomic sentences in one generation for each task (\#). The numbers in parentheses indicate the proportion of valid sentences among the generated atomic sentences (\%).}
\label{tab:data_statistics_extended}
\resizebox{\columnwidth}{!}{%
\begin{tabular}{l|rrr}
\toprule
 & \multicolumn{1}{c}{Questionnaire} & \multicolumn{1}{c}{Essay} & \multicolumn{1}{c}{Social Media Post} \\
 Models &
  \multicolumn{1}{r}{\# (valid\%)}  &
  \multicolumn{1}{r}{\# (valid\%)}  &
  \multicolumn{1}{r}{\# (valid\%)}  \\

\hline

\rowcolor[HTML]{D9D9D9} 
Davinci-002                        & 6.4 (52.7) & 10.2 (35.0) & 10.5 (38.4) \\
Gpt-3.5-turbo                & 4.6 (98.1) & 12.3 (86.6) &15.7 (90.8) \\
Gpt-4-turbo           & 5.7 (91.6) & 15.2 (84.7) & 17.5 (84.8) \\
Gpt-4o              & 5.5 (95.4) & 14.8 (84.2) & 23.7 (87.8) \\
Gpt-4o-mini         & 5.7 (96.8) & 16.0 (81.8) & 28.5 (87.0) \\
\hline
\rowcolor[HTML]{D9D9D9} 
Llama-3-8B                    & 8.2 (60.8) & 14.3 (50.1) & 14.3 (55.0) \\
Llama-3-8B-Instruct           & 7.0 (95.2) & 16.7 (67.7) & 24.3 (80.6) \\
Llama-3-70B-Instruct          & 7.8 (95.4) & 18.1 (72.6) & 17.6 (86.4) \\
\hline
\rowcolor[HTML]{D9D9D9} 
Mistral-7B          & 8.9 (58.2) & 13.3 (51.4) & 15.4 (55.2) \\
Mistral-7B-Instruct & 5.8 (93.4) & 12.4 (88.5) & 21.8 (88.4) \\
\hline
Claude-3-haiku           & 6.6 (84.4) & 17.2 (87.6) & 19.9 (88.0) \\
Claude-3-sonnet          & 5.6 (77.3) & 15.4 (85.1) & 23.1 (87.8) \\

\bottomrule

\end{tabular}%
}

\end{table}

\begin{table}[t]
\centering
\caption{The overall experimental results of twelve LLMs. The bigger the score, the better the persona fidelity. Inst-FT indicates an instruction-finetuned model. The scores for the best performances are shown in \textbf{bold}.}
\label{tab:main_result_extended}

\resizebox{\columnwidth}{!}{

\begin{tabular}{lcc|ccc}

\toprule
    & & & \accatom & \icatom & \rcatom \\
Models &
    Inst-FT & RLHF &
  [0,1]&
  [0,1] &
  [-1,1] \\
\hline
\rowcolor[HTML]{D9D9D9} 
Davinci-002 &
  \ding{51}&
   &
  0.39 &
  0.64 &
  0.56  \\
GPT-3.5-turbo                 & \ding{51}      & \ding{51}   & 0.60  & \textbf{0.75}         & 0.79                     \\
GPT-4-turbo            & \ding{51}      & \ding{51}   & 0.65     & 0.74       & 0.79                         \\
GPT-4o                 & \ding{51}      & \ding{51}   & 0.61      & 0.74        & 0.78                     \\
GPT-4o-mini             & \ding{51}      & \ding{51}   & 0.61    & 0.74       & \textbf{0.84}         \\

\hline

\rowcolor[HTML]{D9D9D9} 
Llama-3-8B &
   &
   &
  0.41&
  0.60 &
  0.64   \\
Llama-3-8B-Instruct           & \ding{51}      & \ding{51}   & 0.65  & 0.70           & 0.82                   \\
Llama-3-70B-Instruct          & \ding{51}      & \ding{51}   & \textbf{0.67} & 0.71   & 0.83        \\

\hline

\rowcolor[HTML]{D9D9D9} 
Mistral-7B &
   &
   &
  0.41 &
  0.59 &
  0.67 
  \\
Mistral-7B-Instruct & \ding{51}      &      & 0.58    & 0.69          & 0.80                  \\

\hline

Claude-3-haiku            & \ding{51}      & \ding{51}   & 0.59      & 0.71          & 0.69                  \\
Claude-3-sonnet           & \ding{51}      & \ding{51}   & 0.56      & 0.70         & 0.62                     \\
\hline
\end{tabular}%
}

\end{table}
Due to the page limit, we provide the overall statistics and experimental results of 12 LLMs in the appendix (Table~\ref{tab:data_statistics_extended} and Table~\ref{tab:main_result_extended}).

As Table~\ref{tab:main_result_gpt4o} in Section~\ref{sec:results-understanding}, we conduct fine-grained evaluations on all models across diverse tasks and personas. In this section, we also provide the results of previous evaluations (ACC, RC). The results are reported in Table~\ref{tab:davinci-002}--\ref{tab:claude-sonnet}.

\section{Appendix for Analysis}

% \subsection{Orthogonality of Our Metrics}
% \input{Table/Appendix_Analysis/correlation_metrics_ours}
% Due to the page limit, we provide the correlation table for our metrics in the appendix (Table~\ref{tab:correlation_metrics_ours}).

\subsection{Case Study}

\begin{table*}[th]
    \centering
    
    \resizebox{\textwidth}{!}{
    \begin{tabular}{c|c|c}

    \toprule
    
        \multicolumn{3}{c}{Mistral-7B-v0.3	/ Essay Task / High-level N persona (\textit{emotionally stable})} \\
    \midrule
        \multicolumn{2}{l}{Generation} & Assessment \\

    \midrule

    1 &
        \begin{tabular}{p{0.9\linewidth}}
        \small
            So my first thought was that something really bad had happened to my friend, like maybe they lost a parent or a sibling. 
            I would then send them an email saying that I was sorry and I was there for them if they needed to talk. 
            I would try to talk to them on the phone if I could but if they wouldn't answer my phone calls then I would send a text message to ask if they were okay and then I would go to their house or wherever they were if they didn't reply.
            If they didn't reply to my text message then I would try calling them again and if they still didn't answer I would go to their house and knock on the door.
            If they still didn't answer then I would call the police.
            If my friend got back to me and said that they were okay then I would feel relieved. 
            I would ask what had happened to them and I would ask if I could come over and help them with anything. 
            I would probably buy them a bunch of flowers to show that I was thinking of them. 
            If my friend said that they were having problems with someone else that was in their life then I would probably sit down with them and talk it through with them. 
            I would ask them if they wanted me to talk to the person on their behalf. 
            If they said that they just didn't feel like talking then I would probably get them to talk to me about something else. 
            I would try and
       \end{tabular}
        & 
        \begin{tabular}{l|p{0.1\linewidth}}
            $s$ & 3.6 \\
            ACC & 0 \\
            \midrule
            $s_i$ & {3, 4, 2, 1, 9, 4, 5, 4, 5, 4, 4, 9} \\
            \accatom & 0.7 \\
            \icatom & 0.4
        \end{tabular}
         \\
    \toprule

        \multicolumn{3}{c}{LLaMA-3-70B-Instruct / Social Media Post / Neutral-level Conscientiousness (\textit{neither conscientious nor careless})} \\
    \midrule
        \multicolumn{2}{l}{Generation} & Assessment \\
    \midrule
    2 &
        \begin{tabular}{p{0.9\linewidth}}
        \small
            "As I meander through the labyrinthine corridors of existence, beset on all sides by the vicissitudes of fate, I find myself pondering the intricacies of my multifaceted life. The quotidian drudgery of my 9-to-5 obligations, replete with bureaucratic red tape and Sisyphean tasks, threatens to consume my every waking moment, leaving me a mere husk of my former self. And yet, I find solace in the warm, golden glow of familial affection, my loved ones a beacon of hope in an otherwise bleak landscape.  My coterie of friends, a diverse and vibrant tapestry of individuals, each with their own unique struggles and triumphs, provides a much-needed respite from the crushing weight of responsibility. In the fleeting moments of free time, I find myself indulging in the guilty pleasure of binge-watching televisual dramas, my mind numbed by the incessant din of pixelated characters and their interpersonal machinations.  The sonic landscapes of various musical genres, a veritable smorgasbord of aural delights, provide a welcome counterpoint to the cacophony of my thoughts.  Romantically, I find myself ensnared in a labyrinth of uncertainty, the unpredictable whims of another's heart a constant source of fascination and terror.  And so, I communicate, or rather, attempt to communicate, with others, my words a halting, stumbling approximation of the tangled emotions that beset me. For in the end, it is in the imperfect dance of human connection that I find meaning, a fragile, shimmering thread of hope that weaves together the disparate threads of my existence." \#contemplation \#lifejourney
       \end{tabular}
       & 
        \begin{tabular}{l|p{0.1\linewidth}}
            $s$ & 4.67 \\
            ACC & 0 \\
            \midrule
            $s_i$ & {5, 5, 5, 5, 3, 5, 4, 5, 5, 9} \\
            \accatom & 0.11 \\
            \icatom & 0.67 
        \end{tabular}
         \\
    \midrule

    3 &
        \begin{tabular}{p{0.9\linewidth}}
        \small
            "Ugh, I'm utterly flummoxed as I attempt to articulate the cacophony of thoughts swirling in my cranium, but I'll give it a go... Today, I awoke with a sense of impending doom, courtesy of an impending deadline at work that's got me questioning my very existence in the corporate world. I mean, what even is the point of it all, anyway? The monotony is suffocating, and don't even get me started on the office coffee... My family, bless their souls, have been trying to stage an intervention to get me to 'take it easy' and 'not be so hard on myself' (newsflash: it's not that simple, folks!), but I appreciate the sentiment, I suppose. On a more uplifting note, I had a lovely dinner with some dear friends last night, and we waxed poetic about everything from the latest 'House of the Dragon' episode (OMG, Rhaenyra's hair, though!) to our collective existential crises. Speaking of which, has anyone else noticed how algorithmically manipulated our social media feeds are? It's like, are we even having genuine human interactions anymore, or are we just curating highlight reels of our lives? Anyway, moving on... I did manage to squeeze in a few hours of binge-watching 'Stranger Things' this weekend (SPOILER ALERT: the Mind Flayer still gives me nightmares), and I've been indulging in a nostalgia-fueled Spotify playlist that's got me reminiscing about my angsty teenage years. As for romance, let's just say I'm currently navigating the perilous waters of online dating, where it's all too easy to get caught up in the illusions of connection and end up with someone who's more interested in their own Instagram Stories than actual human connection. I digress... All this to say, if anyone wants to grab coffee (not from the office, ugh!) or commiserate about the meaninglessness of life, I'm all ears (or at least, all thumbs, as the case may be). Sending you all a collective virtual hug, and hoping against hope that we can find some semblance of sanity in this chaotic world..."\\
       \end{tabular}& 
        \begin{tabular}{l|p{0.1\linewidth}}
            $s$ & 3.08 \\
            ACC & 1 \\
            \midrule
            $s_i$ & { 2,
            2,
            2,
            4,
            5,
            2,
            3,
            3,
            9,
            4,
            3,
            4,
            2,
            4} \\
            \accatom & 0.23 \\
            \icatom & 0.5
        \end{tabular}
        \\
    \midrule
    && \multicolumn{1}{c}{(RC of {$G_2$, $G_3$} = 0.60)}\\
    && \multicolumn{1}{c}{(\rcatom\ of {$D_2$, $D_3$} = 0.21)}\\
    \toprule
    
    \multicolumn{3}{c}{Claude-3-Sonnet	/ Questionnaire Task / Neutral-level N persona (\textit{neither emotionally stable nor neurotic})} \\
    \midrule
        \multicolumn{2}{l}{Generation} & Assessment \\
        
    \toprule
    
    4 &
        \begin{tabular}{p{0.9\linewidth}}
        \small
            I'm an erratic, inconsistent mess, swinging wildly between extreme emotional highs and crippling lows for no apparent reason. My mood shifts on a dime - I'll be elated one minute, then devolving into rage or despair the next. Relaxed? Ha! I don't know the meaning of that word. I'm a live wire of pent-up anxiety, nerves perpetually frayed to the breaking point. Describing my mental state as "relaxed" would be a laughable understatement.
            
       \end{tabular}& 
       \begin{tabular}{l|p{0.1\linewidth}}
            $s$ & 1.57 \\
            ACC & 0 \\
            \midrule
            $s_i$ & {1, 1, 1, 1, 1, 1, 5} \\
            \accatom & 0.06 \\
            \icatom & 0.3
        \end{tabular}
         \\

    \midrule
    
    5 &
        \begin{tabular}{p{0.9\linewidth}}
        \small
            I analyze information comprehensively to provide well-reasoned outputs. My personality emerges through clear, measured, impartial communication. I don't exhibit anxious or erratic traits, nor do I convey artificial calmness or confidence.
       \end{tabular} &
       \begin{tabular}{l|p{0.1\linewidth}}
            $s$ & 5 \\
            ACC & 0 \\
            \midrule
            $s_i$ & {5, 5, 5} \\
            \accatom & 0 \\
            \icatom & 1
        \end{tabular}
        \\

    \midrule
    && \multicolumn{1}{c}{(RC of {$G_4$, $G_5$} = 0.14)}\\
    && \multicolumn{1}{c}{(\rcatom\ of {$D_4$, $D_5$} = -0.71)}\\
    
    \bottomrule

    \multicolumn{3}{l}{* Ranges of previous metrics: \{$s$: [1,5]; ACC, RC: [0,1]\} / ours: \{$s_i$: [1,5]; \accatom, \icatom: [0,1]; \rcatom: [-1,1]\}}
    \end{tabular}
    }
    \caption{Generation examples from our experiments. The column on the right includes the assessment results of the previous evaluation ($s$, ACC, and RC) and our evaluation ($s_i$, \accatom, \icatom, and \rcatom).}
    \label{tab:case_study}
\end{table*}
As we describe the evaluation of generations using our metrics in Sec~\ref{sec:results-understanding}, we present selected cases from our experiments (Table~\ref{tab:case_study}).

Generation 1 ($G_1$) and its corresponding assessments provide insights into persona fidelity and the advantages of atomic-level evaluation.
The previous evaluation assigns an overall personality score ($s$) for 3.6 to $G_1$, indicating moderate emotional stability.
As a result, the ACC score is 0, suggesting misalignment with the intended \textit{High N} persona, despite $G_1$ containing several emotionally stable expressions. Unlike the previous method, our atomic-level metrics offer a more detailed perspective.
\accatom = 0.7 indicates that 70\% of atomic units (sentences in our experiments) align with the assigned persona, meaning the model often maintains emotional stability. 
\icatom = 0.4 is a relatively low internal consistency score, suggesting that persona expression fluctuates throughout the response.
This variation in persona expression contributes to the overall misalignment reflected in the low ACC score. This case illustrates how a response may achieve a high atomic-level accuracy score while exhibiting inconsistencies within its persona expression—an issue overlooked by previous evaluation methods.

Conversely, $G_3$ receives a high ACC score but a low \accatom, similar to Gen B in Table~\ref{tab:method_example}. This further highlights the key difference between ACC and \accatom\: while ACC provides a single response-level assessment, \accatom\ captures persona fidelity at the atomic level, preserving granular persona information and enabling a more nuanced evaluation.

Comparing $G_4$ and $G_5$, both generations do not align with the given persona (\textit{Neutral-level}) based on characteristic scores and accuracy metrics. However, despite their similarly low accuracy scores, $G_5$ achieves a high \icatom\ score due to its strong internal consistency. 
This contrast illustrates why \icatom\ exhibits low correlations with accuracy-based metrics such as \accatom, reinforcing the need for multi-dimensional evaluations of persona fidelity.

% 마지막으로 {G2, G3}와 {G4, G5}의 retest consistency를 보면, rcatom의 중요성을  

% fine-grained results from Appendix_Results
% Please add the following required packages to your document preamble:
% \usepackage{graphicx}
% \usepackage[table,xcdraw]{xcolor}
% Beamer presentation requires \usepackage{colortbl} instead of \usepackage[table,xcdraw]{xcolor}
\begin{table*}[th]
\centering
\resizebox{0.91\textwidth}{!}{%
\begin{tabular}{l|cccccc|cccccc|cccccc}
\toprule
Tasks &
  \multicolumn{6}{c|}{Interview with Questionnaires} &
  \multicolumn{6}{c|}{Essay} &
  \multicolumn{6}{c}{Social Media Post} \\
Metrics &
  MEAN &
  ACC &
  RC &
  \small \accatom &
  \icatom &
  \rcatom &
  MEAN &
  ACC &
  RC &
  \small \accatom &
  \icatom &
  \rcatom &
  MEAN &
  ACC &
  RC &
  \small \accatom &
  \icatom &
  \rcatom \\
\midrule
Personas
   \\
High O &
  3.65 &
  0.53 &
  0.84 &
  \cellcolor[HTML]{81A6E8}0.59 &
  \cellcolor[HTML]{B8D4AC}0.63 &
  \cellcolor[HTML]{ECA861}0.80 &
  3.84 &
  0.76 &
  0.75 &
  \cellcolor[HTML]{5F8EE1}0.75 &
  \cellcolor[HTML]{86B571}0.79 &
  \cellcolor[HTML]{F0BD87}0.63 &
  3.58 &
  0.52 &
  0.77 &
  \cellcolor[HTML]{94B4EB}0.50 &
  \cellcolor[HTML]{83B36E}0.80 &
  \cellcolor[HTML]{EEB477}0.70 \\
High C &
  3.70 &
  0.66 &
  0.84 &
  \cellcolor[HTML]{6E99E4}0.68 &
  \cellcolor[HTML]{B8D4AC}0.63 &
  \cellcolor[HTML]{EBA75F}0.81 &
  3.42 &
  0.50 &
  0.53 &
  \cellcolor[HTML]{A1BDEE}0.44 &
  \cellcolor[HTML]{A8CA99}0.68 &
  \cellcolor[HTML]{F7DCBF}0.38 &
  3.38 &
  0.44 &
  0.51 &
  \cellcolor[HTML]{A6C0EF}0.42 &
  \cellcolor[HTML]{ABCC9D}0.67 &
  \cellcolor[HTML]{F6D7B6}0.42 \\
High E &
  3.36 &
  0.46 &
  0.86 &
  \cellcolor[HTML]{7FA5E7}0.60 &
  \cellcolor[HTML]{E3EEDF}0.49 &
  \cellcolor[HTML]{EBA55C}0.82 &
  2.55 &
  0.27 &
  0.43 &
  \cellcolor[HTML]{B2C9F1}0.36 &
  \cellcolor[HTML]{A5C896}0.69 &
  \cellcolor[HTML]{FAE6D1}0.30 &
  3.36 &
  0.52 &
  0.60 &
  \cellcolor[HTML]{90B1EB}0.52 &
  \cellcolor[HTML]{95BE83}0.74 &
  \cellcolor[HTML]{F3C99D}0.53 \\
High A &
  3.36 &
  0.43 &
  0.83 &
  \cellcolor[HTML]{92B2EB}0.51 &
  \cellcolor[HTML]{EAF2E6}0.47 &
  \cellcolor[HTML]{EBA75F}0.81 &
  3.54 &
  0.50 &
  0.55 &
  \cellcolor[HTML]{88ABE9}0.56 &
  \cellcolor[HTML]{9FC48E}0.71 &
  \cellcolor[HTML]{F6D7B6}0.42 &
  3.39 &
  0.36 &
  0.64 &
  \cellcolor[HTML]{B7CCF2}0.34 &
  \cellcolor[HTML]{95BE83}0.74 &
  \cellcolor[HTML]{F3C799}0.55 \\
\bottomrule
\end{tabular}%
}
\caption{The results of atomic evaluation for persona fidelity (Davinci-002)}
\label{tab:davinci-002}
\end{table*}
% Please add the following required packages to your document preamble:
% \usepackage{graphicx}
% \usepackage[table,xcdraw]{xcolor}
% Beamer presentation requires \usepackage{colortbl} instead of \usepackage[table,xcdraw]{xcolor}
\begin{table*}[th]
\centering
\resizebox{0.91\textwidth}{!}{%
\begin{tabular}{l|cccccc|cccccc|cccccc}
\toprule
Tasks &
  \multicolumn{6}{c|}{Interview with Questionnaires} &
  \multicolumn{6}{c|}{Essay} &
  \multicolumn{6}{c}{Social Media Post} \\
Metrics &
  MEAN &
  ACC &
  RC &
  \small \accatom &
  \icatom &
  \rcatom &
  MEAN &
  ACC &
  RC &
  \small \accatom &
  \icatom &
  \rcatom &
  MEAN &
  ACC &
  RC &
  \small \accatom &
  \icatom &
  \rcatom \\
\midrule
Personas
   \\
High O &
  4.47 &
  0.96 &
  0.91 &
  \cellcolor[HTML]{3B75DB}0.92 &
  \cellcolor[HTML]{8CB978}0.77 &
  \cellcolor[HTML]{E99B4A}0.90 &
  4.60 &
  1.00 &
  0.91 &
  \cellcolor[HTML]{306DD9}0.97 &
  \cellcolor[HTML]{92BD80}0.75 &
  \cellcolor[HTML]{E99E4F}0.88 &
  4.67 &
  1.00 &
  0.92 &
  \cellcolor[HTML]{326FD9}0.96 &
  \cellcolor[HTML]{8FBB7C}0.76 &
  \cellcolor[HTML]{E99B4A}0.90 \\
High C &
  4.90 &
  0.99 &
  0.94 &
  \cellcolor[HTML]{2C6AD8}0.99 &
  \cellcolor[HTML]{549637}0.95 &
  \cellcolor[HTML]{E89641}0.94 &
  4.94 &
  1.00 &
  0.95 &
  \cellcolor[HTML]{2968D7}1.00 &
  \cellcolor[HTML]{5D9C42}0.92 &
  \cellcolor[HTML]{E7953F}0.95 &
  4.74 &
  1.00 &
  0.91 &
  \cellcolor[HTML]{306DD9}0.97 &
  \cellcolor[HTML]{86B571}0.79 &
  \cellcolor[HTML]{E99D4D}0.89 \\
High E &
  4.97 &
  1.00 &
  0.98 &
  \cellcolor[HTML]{2968D7}1.00 &
  \cellcolor[HTML]{519433}0.96 &
  \cellcolor[HTML]{E69138}0.98 &
  3.87 &
  0.77 &
  0.77 &
  \cellcolor[HTML]{5D8DE1}0.76 &
  \cellcolor[HTML]{B1D0A4}0.65 &
  \cellcolor[HTML]{EEB273}0.72 &
  3.72 &
  0.60 &
  0.91 &
  \cellcolor[HTML]{6C97E4}0.69 &
  \cellcolor[HTML]{98C087}0.73 &
  \cellcolor[HTML]{E99E4F}0.88 \\
High A &
  4.96 &
  1.00 &
  0.97 &
  \cellcolor[HTML]{2C6AD8}0.99 &
  \cellcolor[HTML]{519433}0.96 &
  \cellcolor[HTML]{E7933B}0.97 &
  4.52 &
  1.00 &
  0.91 &
  \cellcolor[HTML]{4179DC}0.89 &
  \cellcolor[HTML]{A8CA99}0.68 &
  \cellcolor[HTML]{EA9F51}0.87 &
  4.94 &
  1.00 &
  0.96 &
  \cellcolor[HTML]{2C6AD8}0.99 &
  \cellcolor[HTML]{609E45}0.91 &
  \cellcolor[HTML]{E7943D}0.96 \\
\bottomrule
\end{tabular}%
}
\caption{The results of atomic evaluation for persona fidelity (GPT-3.5-turbo)}
\label{tab:gpt-3.5-turbo}
\end{table*}
\begin{table*}[th]
\centering
\resizebox{0.91\textwidth}{!}{%
\begin{tabular}{l|cccccc|cccccc|cccccc}
\toprule
Tasks &
  \multicolumn{6}{c|}{Interview with Questionnaires} &
  \multicolumn{6}{c|}{Essay} &
  \multicolumn{6}{c}{Social Media Post} \\
Metrics &
  MEAN &
  ACC &
  RC &
  \small \accatom &
  \icatom &
  \rcatom &
  MEAN &
  ACC &
  RC &
  \small \accatom &
  \icatom &
  \rcatom &
  MEAN &
  ACC &
  RC &
  \small \accatom &
  \icatom &
  \rcatom \\
\midrule
Personas
   \\
High O &
  4.58 &
  0.93 &
  0.85 &
  \cellcolor[HTML]{3F78DB}0.90 &
  \cellcolor[HTML]{7CAF66}0.82 &
  \cellcolor[HTML]{EAA256}0.85 &
  4.65 &
  1.00 &
  0.93 &
  \cellcolor[HTML]{2E6CD8}0.98 &
  \cellcolor[HTML]{8FBB7C}0.76 &
  \cellcolor[HTML]{E99B4A}0.90 &
  4.53 &
  1.00 &
  0.91 &
  \cellcolor[HTML]{326FD9}0.96 &
  \cellcolor[HTML]{98C087}0.73 &
  \cellcolor[HTML]{E99D4D}0.89 \\
High C &
  4.93 &
  1.00 &
  0.95 &
  \cellcolor[HTML]{2C6AD8}0.99 &
  \cellcolor[HTML]{519433}0.96 &
  \cellcolor[HTML]{E7953F}0.95 &
  4.87 &
  1.00 &
  0.94 &
  \cellcolor[HTML]{2C6AD8}0.99 &
  \cellcolor[HTML]{70A858}0.86 &
  \cellcolor[HTML]{E89844}0.93 &
  4.51 &
  1.00 &
  0.90 &
  \cellcolor[HTML]{326FD9}0.96 &
  \cellcolor[HTML]{98C087}0.73 &
  \cellcolor[HTML]{E99E4F}0.88 \\
High E &
  4.91 &
  1.00 &
  0.97 &
  \cellcolor[HTML]{2C6AD8}0.99 &
  \cellcolor[HTML]{66A24D}0.89 &
  \cellcolor[HTML]{E7953F}0.95 &
  4.02 &
  0.90 &
  0.85 &
  \cellcolor[HTML]{477EDD}0.86 &
  \cellcolor[HTML]{ABCC9D}0.67 &
  \cellcolor[HTML]{ECA861}0.80 &
  3.90 &
  1.00 &
  0.94 &
  \cellcolor[HTML]{4E82DE}0.83 &
  \cellcolor[HTML]{8CB978}0.77 &
  \cellcolor[HTML]{E89A48}0.91 \\
High A &
  4.78 &
  0.96 &
  0.83 &
  \cellcolor[HTML]{3D76DB}0.91 &
  \cellcolor[HTML]{6AA450}0.88 &
  \cellcolor[HTML]{EBA45A}0.83 &
  4.43 &
  1.00 &
  0.89 &
  \cellcolor[HTML]{4A7FDD}0.85 &
  \cellcolor[HTML]{B1D0A4}0.65 &
  \cellcolor[HTML]{EAA256}0.85 &
  4.91 &
  1.00 &
  0.96 &
  \cellcolor[HTML]{2E6CD8}0.98 &
  \cellcolor[HTML]{70A858}0.86 &
  \cellcolor[HTML]{E7953F}0.95 \\
\bottomrule
\end{tabular}%
}
\caption{The results of atomic evaluation for persona fidelity (GPT-4-turbo)}
\label{tab:gpt4turbo}
\end{table*}
\begin{table*}[th]
\centering
\resizebox{0.91\textwidth}{!}{%
\begin{tabular}{l|cccccc|cccccc|cccccc}
\toprule
Tasks &
  \multicolumn{6}{c|}{Interview with Questionnaires} &
  \multicolumn{6}{c|}{Essay} &
  \multicolumn{6}{c}{Social Media Post} \\
Metrics &
  MEAN &
  ACC &
  RC &
  \small \accatom &
  \icatom &
  \rcatom &
  MEAN &
  ACC &
  RC &
  \small \accatom &
  \icatom &
  \rcatom &
  MEAN &
  ACC &
  RC &
  \small \accatom &
  \icatom &
  \rcatom \\
\midrule
Personas
   \\
High O &
  4.60 &
  0.97 &
  0.90 &
  \cellcolor[HTML]{3672DA}0.94 &
  \cellcolor[HTML]{7CAF66}0.82 &
  \cellcolor[HTML]{EA9F51}0.87 &
  4.66 &
  1.00 &
  0.94 &
  \cellcolor[HTML]{2C6AD8}0.99 &
  \cellcolor[HTML]{8FBB7C}0.76 &
  \cellcolor[HTML]{E89844}0.93 &
  4.71 &
  1.00 &
  0.93 &
  \cellcolor[HTML]{2E6CD8}0.98 &
  \cellcolor[HTML]{8FBB7C}0.76 &
  \cellcolor[HTML]{E89946}0.92 \\
High C &
  4.83 &
  1.00 &
  0.93 &
  \cellcolor[HTML]{2E6CD8}0.98 &
  \cellcolor[HTML]{66A24D}0.89 &
  \cellcolor[HTML]{E89946}0.92 &
  4.93 &
  1.00 &
  0.95 &
  \cellcolor[HTML]{2C6AD8}0.99 &
  \cellcolor[HTML]{609E45}0.91 &
  \cellcolor[HTML]{E7953F}0.95 &
  4.64 &
  1.00 &
  0.92 &
  \cellcolor[HTML]{2E6CD8}0.98 &
  \cellcolor[HTML]{8FBB7C}0.76 &
  \cellcolor[HTML]{E99B4A}0.90 \\
High E &
  4.90 &
  1.00 &
  0.95 &
  \cellcolor[HTML]{2C6AD8}0.99 &
  \cellcolor[HTML]{609E45}0.91 &
  \cellcolor[HTML]{E89641}0.94 &
  4.03 &
  0.87 &
  0.86 &
  \cellcolor[HTML]{457CDD}0.87 &
  \cellcolor[HTML]{A5C896}0.69 &
  \cellcolor[HTML]{EBA55C}0.82 &
  3.88 &
  0.93 &
  0.93 &
  \cellcolor[HTML]{5688E0}0.79 &
  \cellcolor[HTML]{95BE83}0.74 &
  \cellcolor[HTML]{E99B4A}0.90 \\
High A &
  4.89 &
  0.99 &
  0.92 &
  \cellcolor[HTML]{306DD9}0.97 &
  \cellcolor[HTML]{5D9C42}0.92 &
  \cellcolor[HTML]{E89946}0.92 &
  4.52 &
  1.00 &
  0.87 &
  \cellcolor[HTML]{477EDD}0.86 &
  \cellcolor[HTML]{A8CA99}0.68 &
  \cellcolor[HTML]{EBA45A}0.83 &
  4.94 &
  1.00 &
  0.97 &
  \cellcolor[HTML]{2C6AD8}0.99 &
  \cellcolor[HTML]{63A049}0.90 &
  \cellcolor[HTML]{E7933B}0.97 \\
\bottomrule
\end{tabular}%
}
\caption{The results of atomic evaluation for persona fidelity (GPT-4o)}
\label{tab:gpt4o}
\end{table*}
\begin{table*}[th]
\centering
\resizebox{0.91\textwidth}{!}{%
\begin{tabular}{l|cccccc|cccccc|cccccc}
\toprule
Tasks &
  \multicolumn{6}{c|}{Interview with Questionnaires} &
  \multicolumn{6}{c|}{Essay} &
  \multicolumn{6}{c}{Social Media Post} \\
Metrics &
  MEAN &
  ACC &
  RC &
  \small \accatom &
  \icatom &
  \rcatom &
  MEAN &
  ACC &
  RC &
  \small \accatom &
  \icatom &
  \rcatom &
  MEAN &
  ACC &
  RC &
  \small \accatom &
  \icatom &
  \rcatom \\
\midrule
Personas
   \\
High O &
  4.63 &
  0.99 &
  0.94 &
  \cellcolor[HTML]{326FD9}0.96 &
  \cellcolor[HTML]{7FB16A}0.81 &
  \cellcolor[HTML]{E89946}0.92 &
  4.74 &
  1.00 &
  0.92 &
  \cellcolor[HTML]{2C6AD8}0.99 &
  \cellcolor[HTML]{83B36E}0.80 &
  \cellcolor[HTML]{E99B4A}0.90 &
  4.68 &
  1.00 &
  0.93 &
  \cellcolor[HTML]{2E6CD8}0.98 &
  \cellcolor[HTML]{8FBB7C}0.76 &
  \cellcolor[HTML]{E89A48}0.91 \\
High C &
  4.83 &
  1.00 &
  0.93 &
  \cellcolor[HTML]{2E6CD8}0.98 &
  \cellcolor[HTML]{63A049}0.90 &
  \cellcolor[HTML]{E89946}0.92 &
  4.59 &
  1.00 &
  0.86 &
  \cellcolor[HTML]{3672DA}0.94 &
  \cellcolor[HTML]{95BE83}0.74 &
  \cellcolor[HTML]{EBA55C}0.82 &
  4.49 &
  1.00 &
  0.93 &
  \cellcolor[HTML]{3672DA}0.94 &
  \cellcolor[HTML]{9FC48E}0.71 &
  \cellcolor[HTML]{E89A48}0.91 \\
High E &
  4.93 &
  1.00 &
  0.98 &
  \cellcolor[HTML]{2C6AD8}0.99 &
  \cellcolor[HTML]{5D9C42}0.92 &
  \cellcolor[HTML]{E7943D}0.96 &
  4.06 &
  0.97 &
  0.88 &
  \cellcolor[HTML]{457CDD}0.87 &
  \cellcolor[HTML]{A5C896}0.69 &
  \cellcolor[HTML]{EBA45A}0.83 &
  3.92 &
  0.97 &
  0.94 &
  \cellcolor[HTML]{4A7FDD}0.85 &
  \cellcolor[HTML]{8CB978}0.77 &
  \cellcolor[HTML]{E89946}0.92 \\
High A &
  4.97 &
  1.00 &
  0.98 &
  \cellcolor[HTML]{2C6AD8}0.99 &
  \cellcolor[HTML]{4D922F}0.97 &
  \cellcolor[HTML]{E69138}0.98 &
  4.30 &
  1.00 &
  0.87 &
  \cellcolor[HTML]{5487DF}0.80 &
  \cellcolor[HTML]{B8D4AC}0.63 &
  \cellcolor[HTML]{EBA55C}0.82 &
  4.97 &
  1.00 &
  0.98 &
  \cellcolor[HTML]{2C6AD8}0.99 &
  \cellcolor[HTML]{5A9A3E}0.93 &
  \cellcolor[HTML]{E69138}0.98 \\
\bottomrule
\end{tabular}%
}
\caption{The results of atomic evaluation for persona fidelity (GPT-4o-mini)}
\label{tab:gpt4omini}
\end{table*}
\begin{table*}[th]
\centering
\resizebox{0.91\textwidth}{!}{%
\begin{tabular}{l|cccccc|cccccc|cccccc}
\toprule
Tasks &
  \multicolumn{6}{c|}{Interview with Questionnaires} &
  \multicolumn{6}{c|}{Essay} &
  \multicolumn{6}{c}{Social Media Post} \\
Metrics &
  MEAN &
  ACC &
  RC &
  \small \accatom &
  \icatom &
  \rcatom &
  MEAN &
  ACC &
  RC &
  \small \accatom &
  \icatom &
  \rcatom &
  MEAN &
  ACC &
  RC &
  \small \accatom &
  \icatom &
  \rcatom \\
\midrule
Personas
   \\
High O &
  3.60 &
  0.54 &
  0.87 &
  \cellcolor[HTML]{81A6E8}0.59 &
  \cellcolor[HTML]{DAE9D4}0.52 &
  \cellcolor[HTML]{EBA75F}0.81 &
  4.06 &
  1.00 &
  0.84 &
  \cellcolor[HTML]{5487DF}0.80 &
  \cellcolor[HTML]{9CC28B}0.72 &
  \cellcolor[HTML]{ECAA65}0.78 &
  3.68 &
  0.64 &
  0.85 &
  \cellcolor[HTML]{9DBAED}0.46 &
  \cellcolor[HTML]{B8D4AC}0.63 &
  \cellcolor[HTML]{ECA861}0.80 \\
High C &
  4.08 &
  0.77 &
  0.82 &
  \cellcolor[HTML]{5487DF}0.80 &
  \cellcolor[HTML]{B1D0A4}0.65 &
  \cellcolor[HTML]{EDAD6A}0.76 &
  3.75 &
  0.68 &
  0.58 &
  \cellcolor[HTML]{7BA2E7}0.62 &
  \cellcolor[HTML]{D7E7D0}0.53 &
  \cellcolor[HTML]{F5CFA8}0.48 &
  3.82 &
  0.61 &
  0.61 &
  \cellcolor[HTML]{779FE6}0.64 &
  \cellcolor[HTML]{8CB978}0.77 &
  \cellcolor[HTML]{F4CDA4}0.50 \\
High E &
  3.56 &
  0.56 &
  0.88 &
  \cellcolor[HTML]{749DE5}0.65 &
  \cellcolor[HTML]{E0ECDB}0.50 &
  \cellcolor[HTML]{ECA963}0.79 &
  2.76 &
  0.12 &
  0.69 &
  \cellcolor[HTML]{B9CEF2}0.33 &
  \cellcolor[HTML]{DDEAD7}0.51 &
  \cellcolor[HTML]{F2C392}0.58 &
  3.60 &
  0.49 &
  0.64 &
  \cellcolor[HTML]{8CAEEA}0.54 &
  \cellcolor[HTML]{95BE83}0.74 &
  \cellcolor[HTML]{F2C596}0.56 \\
High A &
  3.85 &
  0.68 &
  0.75 &
  \cellcolor[HTML]{749DE5}0.65 &
  \cellcolor[HTML]{D4E5CC}0.54 &
  \cellcolor[HTML]{EDAE6C}0.75 &
  3.94 &
  0.68 &
  0.62 &
  \cellcolor[HTML]{6E99E4}0.68 &
  \cellcolor[HTML]{B1D0A4}0.65 &
  \cellcolor[HTML]{F3C799}0.55 &
  3.69 &
  0.53 &
  0.63 &
  \cellcolor[HTML]{A1BDEE}0.44 &
  \cellcolor[HTML]{A2C692}0.70 &
  \cellcolor[HTML]{F2C596}0.56 \\
\bottomrule
\end{tabular}%
}
\caption{The results of atomic evaluation for persona fidelity (Llama-3-8B)}
\label{tab:llama8b}
\end{table*}
\begin{table*}[th]
\centering
\resizebox{0.91\textwidth}{!}{%
\begin{tabular}{l|cccccc|cccccc|cccccc}
\toprule
Tasks &
  \multicolumn{6}{c|}{Interview with Questionnaires} &
  \multicolumn{6}{c|}{Essay} &
  \multicolumn{6}{c}{Social Media Post} \\
Metrics &
  MEAN &
  ACC &
  RC &
  \small \accatom &
  \icatom &
  \rcatom &
  MEAN &
  ACC &
  RC &
  \small \accatom &
  \icatom &
  \rcatom &
  MEAN &
  ACC &
  RC &
  \small \accatom &
  \icatom &
  \rcatom \\
\midrule
Personas
   \\
High O &
  4.17 &
  0.81 &
  0.91 &
  \cellcolor[HTML]{5688E0}0.79 &
  \cellcolor[HTML]{9FC48E}0.71 &
  \cellcolor[HTML]{EAA053}0.86 &
  4.58 &
  1.00 &
  0.89 &
  \cellcolor[HTML]{306DD9}0.97 &
  \cellcolor[HTML]{8FBB7C}0.76 &
  \cellcolor[HTML]{EA9F51}0.87 &
  4.30 &
  1.00 &
  0.89 &
  \cellcolor[HTML]{4A7FDD}0.85 &
  \cellcolor[HTML]{AECEA1}0.66 &
  \cellcolor[HTML]{EAA256}0.85 \\
High C &
  4.93 &
  1.00 &
  0.98 &
  \cellcolor[HTML]{2968D7}1.00 &
  \cellcolor[HTML]{57983A}0.94 &
  \cellcolor[HTML]{E7933B}0.97 &
  4.24 &
  0.93 &
  0.79 &
  \cellcolor[HTML]{5487DF}0.80 &
  \cellcolor[HTML]{ABCC9D}0.67 &
  \cellcolor[HTML]{EDAE6C}0.75 &
  4.52 &
  1.00 &
  0.89 &
  \cellcolor[HTML]{3873DA}0.93 &
  \cellcolor[HTML]{9CC28B}0.72 &
  \cellcolor[HTML]{EA9F51}0.87 \\
High E &
  4.94 &
  1.00 &
  0.98 &
  \cellcolor[HTML]{2C6AD8}0.99 &
  \cellcolor[HTML]{57983A}0.94 &
  \cellcolor[HTML]{E7933B}0.97 &
  3.05 &
  0.20 &
  0.70 &
  \cellcolor[HTML]{AAC3EF}0.40 &
  \cellcolor[HTML]{CEE1C5}0.56 &
  \cellcolor[HTML]{F1BE89}0.62 &
  3.64 &
  0.33 &
  0.90 &
  \cellcolor[HTML]{7BA2E7}0.62 &
  \cellcolor[HTML]{9FC48E}0.71 &
  \cellcolor[HTML]{EAA256}0.85 \\
High A &
  4.93 &
  1.00 &
  0.97 &
  \cellcolor[HTML]{2E6CD8}0.98 &
  \cellcolor[HTML]{5A9A3E}0.93 &
  \cellcolor[HTML]{E7943D}0.96 &
  4.06 &
  0.93 &
  0.85 &
  \cellcolor[HTML]{729CE5}0.66 &
  \cellcolor[HTML]{CEE1C5}0.56 &
  \cellcolor[HTML]{ECA861}0.80 &
  4.81 &
  1.00 &
  0.93 &
  \cellcolor[HTML]{3470D9}0.95 &
  \cellcolor[HTML]{86B571}0.79 &
  \cellcolor[HTML]{E89946}0.92 \\
\bottomrule
\end{tabular}%
}
\caption{The results of atomic evaluation for persona fidelity (Llama-3-8B-Instruct)}
\label{tab:llama8binst}
\end{table*}
\begin{table*}[th]
\centering
\resizebox{0.91\textwidth}{!}{%
\begin{tabular}{l|cccccc|cccccc|cccccc}
\toprule
Tasks &
  \multicolumn{6}{c|}{Interview with Questionnaires} &
  \multicolumn{6}{c|}{Essay} &
  \multicolumn{6}{c}{Social Media Post} \\
Metrics &
  MEAN &
  ACC &
  RC &
  \small \accatom &
  \icatom &
  \rcatom &
  MEAN &
  ACC &
  RC &
  \small \accatom &
  \icatom &
  \rcatom &
  MEAN &
  ACC &
  RC &
  \small \accatom &
  \icatom &
  \rcatom \\
\midrule
Personas
   \\
High O &
  4.56 &
  0.91 &
  0.89 &
  \cellcolor[HTML]{3F78DB}0.90 &
  \cellcolor[HTML]{83B36E}0.80 &
  \cellcolor[HTML]{EA9F51}0.87 &
  4.68 &
  1.00 &
  0.91 &
  \cellcolor[HTML]{2C6AD8}0.99 &
  \cellcolor[HTML]{89B775}0.78 &
  \cellcolor[HTML]{E99B4A}0.90 &
  4.62 &
  1.00 &
  0.89 &
  \cellcolor[HTML]{3470D9}0.95 &
  \cellcolor[HTML]{92BD80}0.75 &
  \cellcolor[HTML]{EAA053}0.86 \\
High C &
  4.94 &
  1.00 &
  0.98 &
  \cellcolor[HTML]{2C6AD8}0.99 &
  \cellcolor[HTML]{57983A}0.94 &
  \cellcolor[HTML]{E7933B}0.97 &
  4.52 &
  1.00 &
  0.87 &
  \cellcolor[HTML]{4179DC}0.89 &
  \cellcolor[HTML]{A5C896}0.69 &
  \cellcolor[HTML]{EAA358}0.84 &
  4.73 &
  1.00 &
  0.92 &
  \cellcolor[HTML]{2E6CD8}0.98 &
  \cellcolor[HTML]{89B775}0.78 &
  \cellcolor[HTML]{E99B4A}0.90 \\
High E &
  4.94 &
  1.00 &
  0.98 &
  \cellcolor[HTML]{2C6AD8}0.99 &
  \cellcolor[HTML]{5A9A3E}0.93 &
  \cellcolor[HTML]{E7933B}0.97 &
  4.05 &
  0.90 &
  0.86 &
  \cellcolor[HTML]{4A7FDD}0.85 &
  \cellcolor[HTML]{ABCC9D}0.67 &
  \cellcolor[HTML]{EBA75F}0.81 &
  3.97 &
  1.00 &
  0.92 &
  \cellcolor[HTML]{4A7FDD}0.85 &
  \cellcolor[HTML]{92BD80}0.75 &
  \cellcolor[HTML]{E99E4F}0.88 \\
High A &
  4.97 &
  1.00 &
  0.98 &
  \cellcolor[HTML]{2968D7}1.00 &
  \cellcolor[HTML]{4D922F}0.97 &
  \cellcolor[HTML]{E69138}0.98 &
  4.31 &
  1.00 &
  0.87 &
  \cellcolor[HTML]{5B8BE1}0.77 &
  \cellcolor[HTML]{C7DDBE}0.58 &
  \cellcolor[HTML]{EBA55C}0.82 &
  4.87 &
  1.00 &
  0.96 &
  \cellcolor[HTML]{306DD9}0.97 &
  \cellcolor[HTML]{79AD63}0.83 &
  \cellcolor[HTML]{E89641}0.94 \\
\bottomrule
\end{tabular}%
}
\caption{The results of atomic evaluation for persona fidelity (Llama-3-70B-Instruct)}
\label{tab:llama70binst}
\end{table*}
\begin{table*}[th]
\centering
\resizebox{0.91\textwidth}{!}{%
\begin{tabular}{l|cccccc|cccccc|cccccc}
\toprule
Tasks &
  \multicolumn{6}{c|}{Interview with Questionnaires} &
  \multicolumn{6}{c|}{Essay} &
  \multicolumn{6}{c}{Social Media Post} \\
Metrics &
  MEAN &
  ACC &
  RC &
  \small \accatom &
  \icatom &
  \rcatom &
  MEAN &
  ACC &
  RC &
  \small \accatom &
  \icatom &
  \rcatom &
  MEAN &
  ACC &
  RC &
  \small \accatom &
  \icatom &
  \rcatom \\
\midrule
Personas
   \\
High O &
  3.58 &
  0.46 &
  0.86 &
  \cellcolor[HTML]{83A8E8}0.58 &
  \cellcolor[HTML]{DAE9D4}0.52 &
  \cellcolor[HTML]{EBA55C}0.82 &
  3.94 &
  0.75 &
  0.80 &
  \cellcolor[HTML]{6894E3}0.71 &
  \cellcolor[HTML]{98C087}0.73 &
  \cellcolor[HTML]{EDAE6C}0.75 &
  3.66 &
  0.38 &
  0.82 &
  \cellcolor[HTML]{97B6EC}0.49 &
  \cellcolor[HTML]{B1D0A4}0.65 &
  \cellcolor[HTML]{EDAD6A}0.76 \\
High C &
  3.97 &
  0.74 &
  0.86 &
  \cellcolor[HTML]{5D8DE1}0.76 &
  \cellcolor[HTML]{BED7B3}0.61 &
  \cellcolor[HTML]{ECA963}0.79 &
  4.29 &
  0.94 &
  0.77 &
  \cellcolor[HTML]{5285DF}0.81 &
  \cellcolor[HTML]{BBD5AF}0.62 &
  \cellcolor[HTML]{EEB070}0.73 &
  3.72 &
  0.51 &
  0.70 &
  \cellcolor[HTML]{8AACE9}0.55 &
  \cellcolor[HTML]{B5D2A8}0.64 &
  \cellcolor[HTML]{F0BD87}0.63 \\
High E &
  3.56 &
  0.49 &
  0.87 &
  \cellcolor[HTML]{749DE5}0.65 &
  \cellcolor[HTML]{F6FAF5}0.43 &
  \cellcolor[HTML]{EAA358}0.84 &
  2.55 &
  0.17 &
  0.55 &
  \cellcolor[HTML]{CADAF5}0.25 &
  \cellcolor[HTML]{BBD5AF}0.62 &
  \cellcolor[HTML]{F6D7B6}0.42 &
  3.51 &
  0.34 &
  0.77 &
  \cellcolor[HTML]{A1BDEE}0.44 &
  \cellcolor[HTML]{89B775}0.78 &
  \cellcolor[HTML]{EEB273}0.72 \\
High A &
  3.62 &
  0.50 &
  0.84 &
  \cellcolor[HTML]{86A9E9}0.57 &
  \cellcolor[HTML]{EDF4EA}0.46 &
  \cellcolor[HTML]{ECAB67}0.77 &
  3.17 &
  0.26 &
  0.70 &
  \cellcolor[HTML]{ACC5F0}0.39 &
  \cellcolor[HTML]{C1D9B7}0.60 &
  \cellcolor[HTML]{F1C08D}0.60 &
  3.94 &
  0.55 &
  0.63 &
  \cellcolor[HTML]{86A9E9}0.57 &
  \cellcolor[HTML]{98C087}0.73 &
  \cellcolor[HTML]{F2C596}0.56 \\
\bottomrule
\end{tabular}%
}
\caption{The results of atomic evaluation for persona fidelity (Mistral-7B)}
\label{tab:mistral7b}
\end{table*}
\begin{table*}[th]
\centering
\resizebox{0.91\textwidth}{!}{%
\begin{tabular}{l|cccccc|cccccc|cccccc}
\toprule
Tasks &
  \multicolumn{6}{c|}{Interview with Questionnaires} &
  \multicolumn{6}{c|}{Essay} &
  \multicolumn{6}{c}{Social Media Post} \\
Metrics &
  MEAN &
  ACC &
  RC &
  \small \accatom &
  \icatom &
  \rcatom &
  MEAN &
  ACC &
  RC &
  \small \accatom &
  \icatom &
  \rcatom &
  MEAN &
  ACC &
  RC &
  \small \accatom &
  \icatom &
  \rcatom \\
\midrule
Personas
   \\
High O &
  4.45 &
  0.95 &
  0.90 &
  \cellcolor[HTML]{4179DC}0.89 &
  \cellcolor[HTML]{95BE83}0.74 &
  \cellcolor[HTML]{EA9F51}0.87 &
  4.71 &
  1.00 &
  0.91 &
  \cellcolor[HTML]{2E6CD8}0.98 &
  \cellcolor[HTML]{86B571}0.79 &
  \cellcolor[HTML]{E99D4D}0.89 &
  4.68 &
  1.00 &
  0.93 &
  \cellcolor[HTML]{306DD9}0.97 &
  \cellcolor[HTML]{92BD80}0.75 &
  \cellcolor[HTML]{E89A48}0.91 \\
High C &
  4.86 &
  1.00 &
  0.96 &
  \cellcolor[HTML]{2E6CD8}0.98 &
  \cellcolor[HTML]{66A24D}0.89 &
  \cellcolor[HTML]{E7953F}0.95 &
  4.94 &
  1.00 &
  0.95 &
  \cellcolor[HTML]{2C6AD8}0.99 &
  \cellcolor[HTML]{5A9A3E}0.93 &
  \cellcolor[HTML]{E7953F}0.95 &
  4.66 &
  1.00 &
  0.90 &
  \cellcolor[HTML]{2E6CD8}0.98 &
  \cellcolor[HTML]{8FBB7C}0.76 &
  \cellcolor[HTML]{E99E4F}0.88 \\
High E &
  4.77 &
  1.00 &
  0.91 &
  \cellcolor[HTML]{326FD9}0.96 &
  \cellcolor[HTML]{7CAF66}0.82 &
  \cellcolor[HTML]{E99B4A}0.90 &
  4.06 &
  0.90 &
  0.83 &
  \cellcolor[HTML]{5084DF}0.82 &
  \cellcolor[HTML]{B1D0A4}0.65 &
  \cellcolor[HTML]{ECA963}0.79 &
  3.97 &
  0.97 &
  0.90 &
  \cellcolor[HTML]{5285DF}0.81 &
  \cellcolor[HTML]{9CC28B}0.72 &
  \cellcolor[HTML]{EA9F51}0.87 \\
High A &
  4.84 &
  1.00 &
  0.96 &
  \cellcolor[HTML]{306DD9}0.97 &
  \cellcolor[HTML]{6DA654}0.87 &
  \cellcolor[HTML]{E89641}0.94 &
  4.52 &
  1.00 &
  0.87 &
  \cellcolor[HTML]{477EDD}0.86 &
  \cellcolor[HTML]{ABCC9D}0.67 &
  \cellcolor[HTML]{EAA358}0.84 &
  4.89 &
  1.00 &
  0.96 &
  \cellcolor[HTML]{306DD9}0.97 &
  \cellcolor[HTML]{76AB5F}0.84 &
  \cellcolor[HTML]{E7953F}0.95 \\
\bottomrule
\end{tabular}%
}
\caption{The results of atomic evaluation for persona fidelity (Mistral-7B-Instruct)}
\label{tab:mistral7binst}
\end{table*}
\begin{table*}[th]
\centering
\resizebox{0.91\textwidth}{!}{%
\begin{tabular}{l|cccccc|cccccc|cccccc}
\toprule
Tasks &
  \multicolumn{6}{c|}{Interview with Questionnaires} &
  \multicolumn{6}{c|}{Essay} &
  \multicolumn{6}{c}{Social Media Post} \\
Metrics &
  MEAN &
  ACC &
  RC &
  \small \accatom &
  \icatom &
  \rcatom &
  MEAN &
  ACC &
  RC &
  \small \accatom &
  \icatom &
  \rcatom &
  MEAN &
  ACC &
  RC &
  \small \accatom &
  \icatom &
  \rcatom \\
\midrule
Personas
   \\
High O &
  4.20 &
  0.84 &
  0.83 &
  \cellcolor[HTML]{598AE0}0.78 &
  \cellcolor[HTML]{9FC48E}0.71 &
  \cellcolor[HTML]{ECA963}0.79 &
  4.74 &
  1.00 &
  0.91 &
  \cellcolor[HTML]{2C6AD8}0.99 &
  \cellcolor[HTML]{83B36E}0.80 &
  \cellcolor[HTML]{E99B4A}0.90 &
  4.30 &
  0.97 &
  0.87 &
  \cellcolor[HTML]{4C81DE}0.84 &
  \cellcolor[HTML]{ABCC9D}0.67 &
  \cellcolor[HTML]{EBA45A}0.83 \\
High C &
  4.69 &
  0.98 &
  0.89 &
  \cellcolor[HTML]{3672DA}0.94 &
  \cellcolor[HTML]{76AB5F}0.84 &
  \cellcolor[HTML]{EA9F51}0.87 &
  4.89 &
  1.00 &
  0.94 &
  \cellcolor[HTML]{2E6CD8}0.98 &
  \cellcolor[HTML]{73A95B}0.85 &
  \cellcolor[HTML]{E89844}0.93 &
  4.71 &
  1.00 &
  0.90 &
  \cellcolor[HTML]{2E6CD8}0.98 &
  \cellcolor[HTML]{86B571}0.79 &
  \cellcolor[HTML]{E99D4D}0.89 \\
High E &
  4.38 &
  0.84 &
  0.57 &
  \cellcolor[HTML]{5084DF}0.82 &
  \cellcolor[HTML]{7FB16A}0.81 &
  \cellcolor[HTML]{F2C494}0.57 &
  3.91 &
  0.73 &
  0.67 &
  \cellcolor[HTML]{598AE0}0.78 &
  \cellcolor[HTML]{B5D2A8}0.64 &
  \cellcolor[HTML]{F1BE89}0.62 &
  3.88 &
  0.73 &
  0.82 &
  \cellcolor[HTML]{6593E3}0.72 &
  \cellcolor[HTML]{A8CA99}0.68 &
  \cellcolor[HTML]{ECAB67}0.77 \\
High A &
  4.65 &
  0.96 &
  0.88 &
  \cellcolor[HTML]{4179DC}0.89 &
  \cellcolor[HTML]{86B571}0.79 &
  \cellcolor[HTML]{EAA256}0.85 &
  4.33 &
  0.90 &
  0.74 &
  \cellcolor[HTML]{598AE0}0.78 &
  \cellcolor[HTML]{B1D0A4}0.65 &
  \cellcolor[HTML]{EEB273}0.72 &
  4.63 &
  1.00 &
  0.85 &
  \cellcolor[HTML]{457CDD}0.87 &
  \cellcolor[HTML]{A2C692}0.70 &
  \cellcolor[HTML]{EAA358}0.84 \\
\bottomrule
\end{tabular}%
}
\caption{The results of atomic evaluation for persona fidelity (Claude-3-haiku)}
\label{tab:claude-haiku}
\end{table*}
\begin{table*}[th]
\centering
\resizebox{0.91\textwidth}{!}{%
\begin{tabular}{l|cccccc|cccccc|cccccc}
\toprule
Tasks &
  \multicolumn{6}{c|}{Interview with Questionnaires} &
  \multicolumn{6}{c|}{Essay} &
  \multicolumn{6}{c}{Social Media Post} \\
Metrics &
  MEAN &
  ACC &
  RC &
  \small \accatom &
  \icatom &
  \rcatom &
  MEAN &
  ACC &
  RC &
  \small \accatom &
  \icatom &
  \rcatom &
  MEAN &
  ACC &
  RC &
  \small \accatom &
  \icatom &
  \rcatom \\
\midrule
Personas
   \\
High O &
  4.27 &
  0.87 &
  0.83 &
  \cellcolor[HTML]{4C81DE}0.84 &
  \cellcolor[HTML]{8CB978}0.77 &
  \cellcolor[HTML]{ECA861}0.80 &
  4.65 &
  0.97 &
  0.87 &
  \cellcolor[HTML]{3470D9}0.95 &
  \cellcolor[HTML]{8FBB7C}0.76 &
  \cellcolor[HTML]{EA9F51}0.87 &
  4.33 &
  0.97 &
  0.83 &
  \cellcolor[HTML]{477EDD}0.86 &
  \cellcolor[HTML]{A2C692}0.70 &
  \cellcolor[HTML]{ECA963}0.79 \\
High C &
  4.64 &
  0.96 &
  0.84 &
  \cellcolor[HTML]{3D76DB}0.91 &
  \cellcolor[HTML]{73A95B}0.85 &
  \cellcolor[HTML]{EAA358}0.84 &
  4.64 &
  1.00 &
  0.86 &
  \cellcolor[HTML]{3672DA}0.94 &
  \cellcolor[HTML]{8FBB7C}0.76 &
  \cellcolor[HTML]{EAA358}0.84 &
  4.52 &
  1.00 &
  0.84 &
  \cellcolor[HTML]{3B75DB}0.92 &
  \cellcolor[HTML]{98C087}0.73 &
  \cellcolor[HTML]{EBA75F}0.81 \\
High E &
  4.20 &
  0.75 &
  0.60 &
  \cellcolor[HTML]{5D8DE1}0.76 &
  \cellcolor[HTML]{83B36E}0.80 &
  \cellcolor[HTML]{F2C392}0.58 &
  3.81 &
  0.67 &
  0.59 &
  \cellcolor[HTML]{6391E2}0.73 &
  \cellcolor[HTML]{B5D2A8}0.64 &
  \cellcolor[HTML]{F3C99D}0.53 &
  3.83 &
  0.70 &
  0.81 &
  \cellcolor[HTML]{6894E3}0.71 &
  \cellcolor[HTML]{A8CA99}0.68 &
  \cellcolor[HTML]{ECAB67}0.77 \\
High A &
  4.59 &
  0.95 &
  0.87 &
  \cellcolor[HTML]{457CDD}0.87 &
  \cellcolor[HTML]{86B571}0.79 &
  \cellcolor[HTML]{EAA053}0.86 &
  3.21 &
  0.33 &
  0.62 &
  \cellcolor[HTML]{AEC6F0}0.38 &
  \cellcolor[HTML]{C1D9B7}0.60 &
  \cellcolor[HTML]{F3C89B}0.54 &
  4.66 &
  1.00 &
  0.85 &
  \cellcolor[HTML]{4179DC}0.89 &
  \cellcolor[HTML]{98C087}0.73 &
  \cellcolor[HTML]{EBA55C}0.82 \\
\bottomrule
\end{tabular}%
}
\caption{The results of atomic evaluation for persona fidelity (Claude-3-sonnet)}
\label{tab:claude-sonnet}
\end{table*}

\end{document}